\documentclass{article} 
\usepackage[final]{colm2026_conference}

\usepackage{microtype}
\usepackage{hyperref}
\usepackage{url}
\usepackage{booktabs}

\usepackage{lineno}

\usepackage{graphicx}
\usepackage{amsmath}
\usepackage{subcaption}
\usepackage{tabularx}
\usepackage{array}

\usepackage{cleveref}
\crefname{equation}{Equation}{Equations}
\crefname{figure}{Figure}{Figures}
\crefname{table}{Table}{Tables}
\crefname{section}{\S}{\S}
\crefname{subfigure}{Figure}{Figures}

\definecolor{darkblue}{rgb}{0, 0, 0.5}
\hypersetup{colorlinks=true, citecolor=darkblue, linkcolor=darkblue, urlcolor=darkblue}

\title{Measuring Task-Agnostic Training Data Influence \\ Across Language Model Pretraining}

\newcommand{\NAIST}{1}
\newcommand{\LLMC}{2}
\newcommand{\UWaseda}{3}
\newcommand{\UTokyo}{4}
\newcommand{\ITU}{5}
\newcommand{\UTohoku}{6}

\newcommand{\affmark}[1]{%
  \textsuperscript{\normalfont\mdseries #1}%
}

\author{%
\textbf{Yuto Nishida}\affmark{\NAIST,\LLMC} \quad
\textbf{Hirokazu Kiyomaru}\affmark{\LLMC} \quad
\textbf{Yusuke Oda}\affmark{\LLMC} \quad
\textbf{Takashi Kodama}\affmark{\LLMC}
\\
\textbf{Chaoran Liu}\affmark{\LLMC} \quad
\textbf{Daisuke Kawahara}\affmark{\LLMC,\UWaseda} \quad
\textbf{Yusuke Miyao}\affmark{\LLMC,\UTokyo} \\
\textbf{Max Müller-Eberstein}\affmark{\UTokyo,\ITU} \quad
\textbf{Masaru Isonuma}\affmark{\LLMC,\UTohoku}
\\[0.5em]
\affmark{\NAIST}Nara Institute of Science and Technology \quad
\affmark{\LLMC}NII LLMC \quad
\affmark{\UWaseda}Waseda University
\\
\affmark{\UTokyo}The University of Tokyo \quad
\affmark{\ITU}IT University of Copenhagen \quad
\affmark{\UTohoku}Tohoku University
\\[0.4em]
\textnormal{Contact: }\texttt{nishida.yuto.nu8@is.naist.jp}
}

\begin{document}

\ifcolmsubmission
\linenumbers
\fi

\maketitle

\begin{abstract}
Measuring training data influence consistently across language model pretraining is challenging.
It is difficult to select downstream tasks or validation sets representative of a model's general capabilities, and reliance on task performance at intermediate checkpoints complicates comparisons across training.
We propose a measure of training data influence that does not require selecting a downstream task or validation set as the attribution target.
Specifically, we define an example's influence by how much its gradient update reduces the squared distance to the final parameters of a given pretraining run, and estimate this quantity from intermediate checkpoints without retraining.
Applying the method to 18 configurations from the Pythia and PolyPythia suites, we find systematic temporal changes in influential data.
Early in training, literature-related data are more strongly aligned with the trajectory toward the final parameters, whereas STEM data become more strongly aligned in later stages.
This qualitative crossover is broadly consistent across model configurations.
Our results provide a tractable trajectory-level view of how influential data change throughout pretraining, complementing influence analyses defined with respect to specific downstream tasks or validation sets.
\end{abstract}

\section{Introduction}
\label{sec:introduction}

Understanding how individual training examples shape a model has long been an important problem in machine learning~\citep{pmlr-v70-koh17a,pmlr-v97-ghorbani19c,pmlr-v162-ilyas22a}.
Such analyses are not only useful for answering questions about what a model learns from its data, but also for practical goals such as debugging anomalous predictions~\citep{pmlr-v70-koh17a}, identifying harmful or biased training instances~\citep{pan2025detecting,jiao2025datelm}, guiding data curation~\citep{yu2024mates,xia2024less}, tracing data provenance~\citep{akyurek-etal-2022-towards}, and supporting data valuation~\citep{pmlr-v97-ghorbani19c,choe2025what}.

Most existing methods for training data influence define an example's effect with respect to a specific target behavior, such as the loss or prediction on a test point, or downstream task performance on an evaluation metric~\citep{pmlr-v70-koh17a,pmlr-v97-ghorbani19c,pruthi2020tracin,hammoudeh2024survey}.
Such task-specific analyses are natural when the goal is to explain or improve performance on a particular task and have already revealed informative patterns in pretraining, including shifts in which domains matter most for particular capabilities~\citep{rosa-etal-2025-impact,wang2025data}.

Language model pretraining, however, aims to develop broad, general capabilities rather than optimize for a single task or capability.
Therefore, it is difficult to select downstream tasks or validation sets that are representative of these general capabilities throughout training.
Moreover, influence estimates tied to task performance at intermediate checkpoints can be difficult to compare consistently across the full pretraining trajectory, since early checkpoints may have acquired capabilities relevant to the target task without yet being able to fully solve it.

\begin{figure}[t]
\centering
\includegraphics[width=1.0\linewidth]{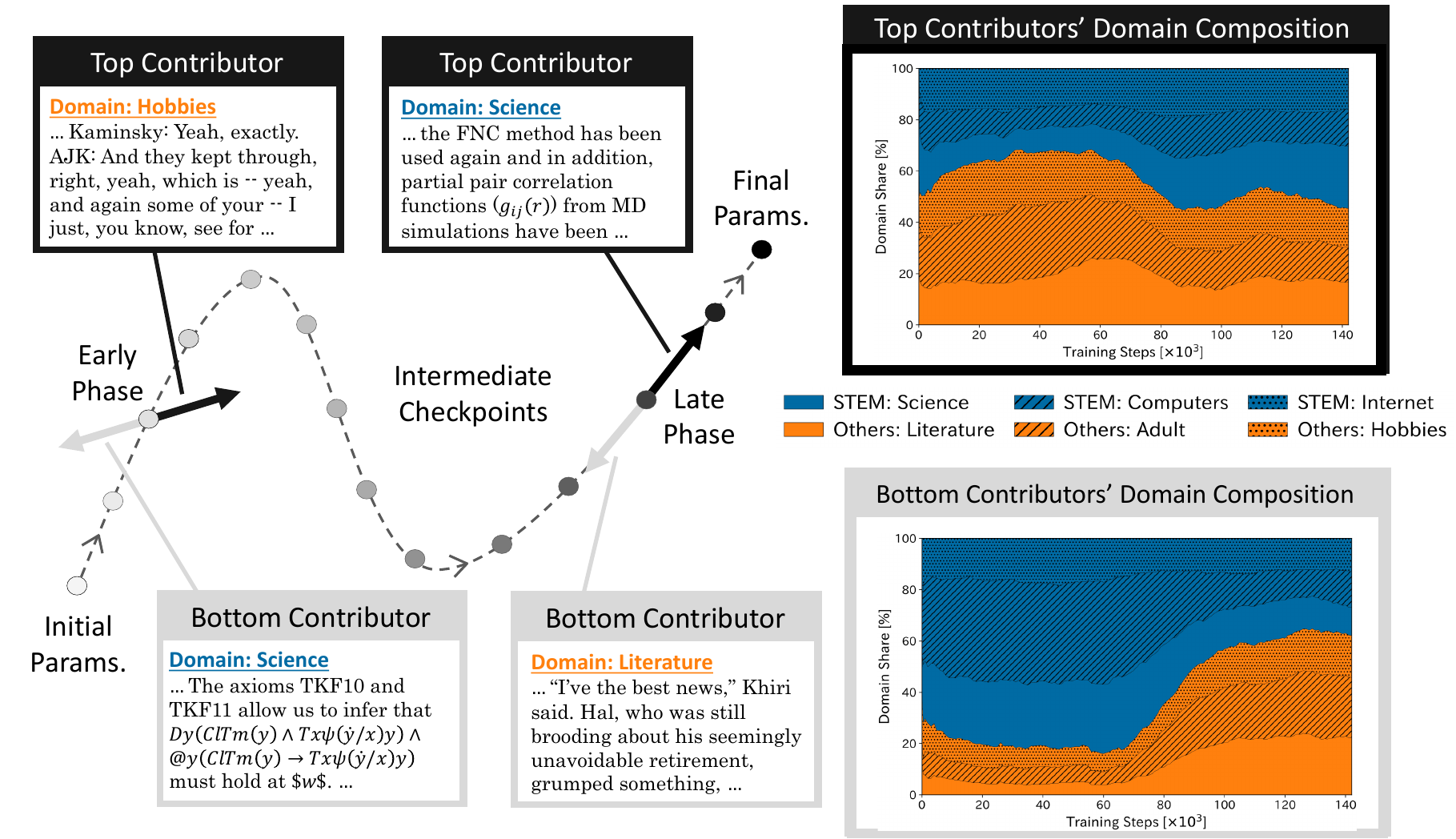}
\caption{Overview of our task-agnostic, training data contribution measure: Top contributors move the model towards its final parameters, while bottom contributors yield negligible or opposing updates. Across pretraining, our experiments reveal a cross-over of STEM versus non-STEM domains, in which the former gain importance during later training.}
\label{fig:overview}
\end{figure}

In this work, we reformulate training data influence across language model pretraining in task-agnostic terms.
Rather than tying influence to a downstream metric, we view it as the extent to which an individual training example moves the model toward the final parameters of a given pretraining run (\cref{fig:overview}).
Specifically, we quantify an example's influence by how much its gradient update reduces the squared L2 distance to the final model weights, and estimate this quantity from intermediate checkpoints without retraining.
The final model thus serves as a common reference point throughout pretraining, yielding a notion of influence that can be compared consistently across different stages of training without relying on task labels or benchmark accuracy.

Across the full pretraining trajectories of 18 variants of the Pythia model family~\citep{pmlr-v202-biderman23a,wal2025polypythias}, each with 154 checkpoints, we find consistent stage-dependent shifts in which kinds of data are most influential.
For instance, literature data are more influential during early training, whereas STEM data become increasingly influential in later stages. 
These results suggest that the training examples that most strongly move the model toward its final pretrained parameters are not static across pretraining, motivating stage-aware approaches to pretraining data composition and curriculum design.

Our contributions are threefold:

\begin{itemize}
    \item We reformulate training data influence across language model pretraining in task-agnostic terms and derive a checkpoint-based approximation for post hoc estimation without retraining (\cref{sec:method}).
    \item We characterize how training data contribution evolves throughout pretraining, revealing systematic stage-dependent variation across different types of training data (\cref{sec:experiments}).
    \item We establish the reliability and scope of the proposed analysis across checkpoint approximation, reference endpoints, model configurations, and attribution targets (\cref{sec:analysis}).
\end{itemize}

\section{Related Work}
\label{sec_related_work}

Training data influence has been studied from multiple perspectives:
Influence Functions~\citep{pmlr-v70-koh17a} estimate how upweighting or removing a training point changes a test prediction or loss, and Data Shapley~\citep{pmlr-v97-ghorbani19c} assigns each datum a marginal value with respect to predictive utility.
More broadly, prior work typically defines a training example's effect with respect to a specific task, such as a prediction, loss, or performance measure~\citep{hammoudeh2024survey}.
Our work differs in replacing such task-conditioned targets with progress toward the final pretrained model.

Because estimating training data influence directly is often computationally expensive, prior work has developed scalable approximations.
TracIn~\citep{pruthi2020tracin} estimates the influence of a training example on a test prediction by tracing gradients through saved checkpoints.
TRAK~\citep{pmlr-v202-park23c} improves the tractability-efficacy trade-off for large non-convex models.
More recent methods extend influence estimation and data valuation to foundation-model scale, including LoGra~\citep{choe2025what} for scalable influence functions in LLMs.
Our method shares this goal of tractable influence estimation at scale: by leveraging intermediate checkpoints from an existing pretraining run, it can be applied to LLMs without retraining from scratch.

Influence analysis has also been applied to language model pretraining and has begun to uncover systematic patterns in how pretraining data shapes a model.
\citet{wang2025data} show that influence patterns evolve substantially over training: for a math-related validation corpus, specialized domains such as ArXiv and GitHub receive disproportionately high value, while more general web corpora contribute more early in training and diminish later.
\citet{chang2025scalable} study fact tracing in LLM pretraining and find that highly influential examples are not always passages that explicitly state the queried fact, highlighting a gap between causal influence and factual attribution that narrows with scale.
DATE-LM~\citep{jiao2025datelm} further shows that, when data attribution methods are evaluated through applications such as training-data selection, toxicity/bias filtering, and factual attribution, no single method dominates across tasks and performance is sensitive to task-specific evaluation design.
Together, these results suggest that task-conditioned evaluation offers only a partial view of how pretraining data shapes a model, motivating a task-agnostic perspective on training data influence across pretraining.

\section{Method}
\label{sec:method}

In this section, we introduce a task-agnostic measure of training data influence based on reduction in parameter-space distance to the final parameters of a given pretraining run.
Our formulation is inspired by trajectory-matching methods in dataset distillation~\citep{cazenavette2022distillation,pmlr-v202-cui23e,liu2024dataset}, but uses parameter-space geometry retrospectively to quantify how real training examples move the model toward its final parameters rather than to optimize synthetic data.
We first define this quantity at the mini-batch and example levels, and then derive a checkpoint-based approximation that allows it to be estimated post hoc from intermediate checkpoints without retraining.

\subsection{Mini-Batch Influence}
\label{sec:minibatch_influence}

Let updates be indexed by $t = 0,1,\dots,T-1$, let $\theta_t \in \mathbb{R}^d$ denote the model parameters before update $t$, and let $B_t = (x_1^t,\dots,x_N^t)$ denote the mini-batch used at that update.
The corresponding parameter update is $\Delta_t := \theta_{t+1} - \theta_t$. Given the final pretrained parameters $\theta^{*} := \theta_T$, and $S_t := \|\theta^{*} - \theta_t\|_2^2$ as the squared L2 distance from the current parameters to the final model, we define the influence of mini-batch $B_t$ as the reduction in distance to the final parameters induced by the update at step $t$:
\begin{equation}
  \mathrm{Cont}(B_t) := S_t - S_{t+1}.
  \label{eq:cont_batch_def}
\end{equation}
This quantity is exact and independent of optimizer-specific decompositions.
Thus, expanding $S_{t+1}$ using $\theta_{t+1} = \theta_t + \Delta_t$ gives
$S_{t+1} = \|\theta^{*} - \theta_t\|_2^2 - 2\Delta_t^{\top}(\theta^{*} - \theta_t) + \|\Delta_t\|_2^2$, yielding
\begin{equation}
  \mathrm{Cont}(B_t)
  = 2\Delta_t^{\top}(\theta^{*} - \theta_t) - \|\Delta_t\|_2^2.
  \label{eq:cont_batch_closed}
\end{equation}
The first term measures the alignment between the update direction and the direction towards the final parameters.
The second term penalizes the squared norm of the update.
Therefore, a mini-batch has high influence when the alignment term outweighs the squared-norm penalty.
$\mathrm{Cont}(B_t)$ can be negative when the update is poorly aligned with the direction toward the final parameters, indicating that the update moves the model away from the final model in terms of squared distance.

\subsection{Example-Level Influence}
\label{sec:example_influence}

To decompose mini-batch updates into example-level terms, we assume for analysis that optimization is performed by standard stochastic gradient descent (SGD) with loss function $\ell$ and learning rate $\eta_t$ at step $t$.\footnote{
Actual LLM pretraining typically uses adaptive optimizers such as Adam~\citep{kingma2015adam}.
We adopt the SGD assumption only for the local example-level decomposition, following the SGD-based formulation used in TracIn~\citep{pruthi2020tracin}.
Under adaptive optimizers, an example's exact effect on the current update also depends on optimizer states carried over from previous mini-batches, making the decomposition substantially more involved.
Our goal here is to obtain a tractable per-example approximation for influence analysis.
}
The update induced by example $x_k^t$ is $\Delta_{t,k} := -\eta_t \nabla_{\theta}\ell(x_k^t;\theta_t)$, so that $\Delta_t = \sum_{k=1}^{N}\Delta_{t,k}$.
We define the influence of example $x_k^t$ by
\begin{equation}
  \mathrm{Cont}(x_k^t)
  := 2\Delta_{t,k}^{\top}(\theta^{*} - \theta_t) - \Delta_{t,k}^{\top}\Delta_t.
  \label{eq:cont_sample_def}
\end{equation}
By construction, the example-level influences sum to the mini-batch influence, i.e., $\sum_{k=1}^{N}\mathrm{Cont}(x_k^t) = \mathrm{Cont}(B_t)$. This definition also shows that example-level influence depends on other examples in the same mini-batch.
Expanding the second term in \cref{eq:cont_sample_def} gives $\mathrm{Cont}(x_k^t) = 2\Delta_{t,k}^{\top}(\theta^{*} - \theta_t) - |\Delta_{t,k}|2^2 - \sum_{j \neq k}\Delta_{t,k}^{\top}\Delta_{t,j}$, so the penalty decomposes into an example-level norm term and cross-example interaction terms within the mini-batch.

Using the simplifying assumption that per-example update vectors are pairwise orthogonal within the mini-batch, we can see the connection of our method with TracIn~\citep{pruthi2020tracin} in that the interaction term vanishes and the remaining norm term becomes $\|\Delta_{t,k}\|_2^2 = \eta_t^2 \|\nabla_{\theta}\ell(x_k^t;\theta_t)\|_2^2$.
This term corresponds, up to scalar factors determined by the learning rate and batching convention, to the \emph{self-influence} term in TracIn for the special case where the training and test examples coincide.
\citet{pruthi2020tracin} note that incorrectly labeled examples tend to have large self-influence.
Our formulation is consistent with this observation: for potentially noisy examples, such as mislabeled ones, the corresponding self-influence-like norm term tends to be large, so it acts as a penalty for example-level influence, assigning smaller influence to such examples.

\subsection{Checkpoint-Based Approximation}
\label{sec:approx_interval}

Computing the quantities above for every training step would require access to all intermediate states or retraining the model.
In practice, we instead assume access to periodically saved checkpoints from an existing pretraining run. Let $c < c'$ be consecutive checkpoint steps such that $c \le t < c'$, and let $\theta_c$ and $\theta_{c'}$ denote the corresponding saved parameters. We approximate $\theta_t \approx \theta_c$ and $\Delta_t \approx \theta_{c'} - \theta_c$, i.e., we treat all examples in the interval $[c,c')$ as if they were processed at the checkpoint state $\theta_c$. For an example $x_k^t$ in that interval, let $\Delta_{c,k} := -\eta_t \nabla_{\theta}\ell(x_k^t;\theta_c)$.
We then estimate example-level influence by
\begin{equation}
  \mathrm{Cont}(x_k^t)
  \approx
  2\Delta_{c,k}^{\top}(\theta^{*} - \theta_c)
  - \Delta_{c,k}^{\top}(\theta_{c'} - \theta_c).
  \label{eq:cont_approx_interval}
\end{equation}
This approximation replaces step-level quantities by checkpoint-level quantities and allows the proposed measure to be applied post hoc to large pretraining runs without retraining from scratch.
In our experiments, we use \cref{eq:cont_approx_interval} to estimate example-level influence from publicly available checkpoints.

\section{Experiments}
\label{sec:experiments}

We apply the proposed measure to characterize how training data contribution evolves across pretraining and which kinds of data contribute most strongly at different stages.

\subsection{Setup}
\label{sec:experimental_setup}

We evaluate the proposed influence measure on multiple publicly released pretraining trajectories.
For the Pythia suite~\citep{pmlr-v202-biderman23a}, we use the deduplicated models at six scales: 70M, 160M, 410M, 1.4B, 6.9B, and 12B.
Each model is trained on 300B tokens from The Pile~\citep{gao2020pile800gbdatasetdiverse} and provides 154 checkpoints at steps $0, 1, 2, 4, \dots, 512, 1000, 2000, \dots, 143000$.

We additionally analyze the PolyPythias~\citep{wal2025polypythias} at the 160M and 410M scales to examine the robustness of our findings to variations in model initialization and data ordering.
For each model size, we use three runs in which both factors vary.
At the 160M scale, we further use three runs from each of the decoupled \texttt{data-seed} and \texttt{weight-seed} variants, allowing the effects of data ordering and weight initialization to be examined separately.
Each PolyPythia run likewise provides 154 checkpoints.

The Pythia and PolyPythia models are trained with batches of 1024 sequences, each containing 2048 tokens.
We treat each 2048-token sequence as one training example.
In all experiments, we sample examples from the actual training data stream associated with each checkpoint interval, rather than from held-out or proxy corpora.
This allows us to relate the estimated influence directly to the data observed by the model during pretraining.

We estimate example-level influence using the checkpoint-based approximation described in \cref{sec:approx_interval}, namely \cref{eq:cont_approx_interval}.
This allows us to estimate influence post hoc from publicly released checkpoints without retraining the model.
We first examine the overall trajectory of influence in \cref{sec:contribution_trajectory}, and then analyze its relationships with text difficulty (\cref{sec:contribution_difficulty}) and textual domain (\cref{sec:contribution_domain}).
Because the main qualitative findings are broadly consistent across model configurations, we first report results for Pythia-1.4B-Deduped and then examine robustness across model sizes, initializations, and data orderings in \cref{sec:analysis}.

\subsection{Trajectory of the Contribution Distribution}
\label{sec:contribution_trajectory}

\begin{figure*}[t]
\captionsetup[subfigure]{justification=centering, skip=0pt}
  \centering
  \begin{minipage}[b]{0.53\textwidth}
    \centering
    \includegraphics[width=\textwidth]{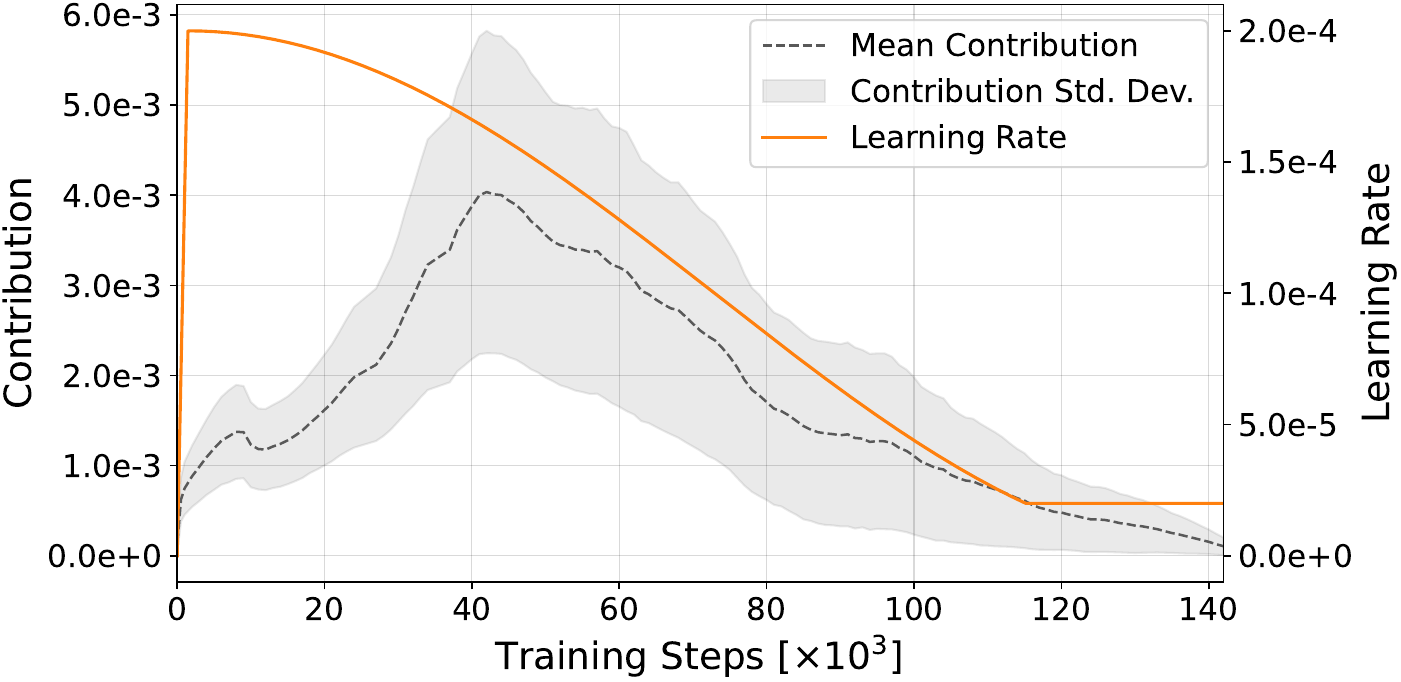}
    \subcaption{Contribution statistics over training}
    \label{fig:contribution_stats}
  \end{minipage}
  \hfill
  \begin{minipage}[b]{0.44\textwidth}
    \centering
    \includegraphics[width=\textwidth]{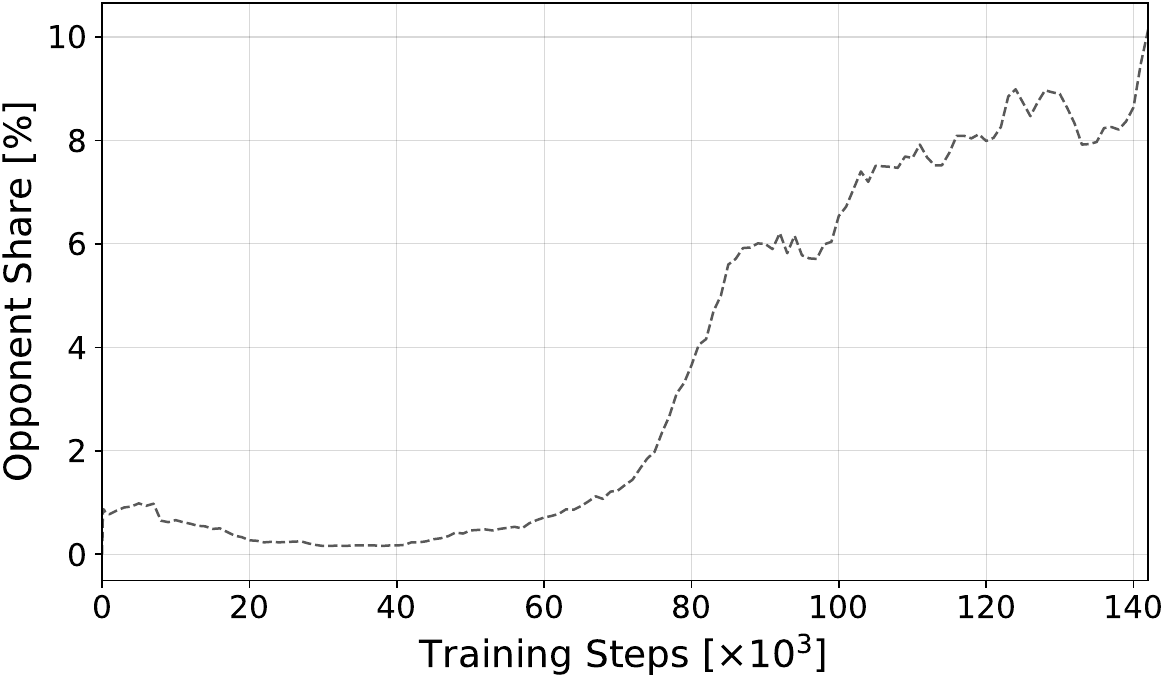}
    \subcaption{Share of opponent examples}
    \label{fig:opponent_share}
  \end{minipage}
  \caption{Training dynamics of the contribution distribution on Pythia-1.4B-Deduped. All curves are moving averages with window size 10.}
  \label{fig:contribution_dynamics}
\end{figure*}

We first characterize how the distribution of contribution, which measures how strongly each example's update moves the model toward its final parameters, evolves over training.
For each checkpoint interval, we uniformly sample 1,000 examples from the actual training data and compute their contributions.
We summarize the distribution using its mean and standard deviation, together with the share of examples with negative contribution.
We refer to the latter as \emph{opponents}, since under our definition their updates move the model away from its final parameters.

As shown in \cref{fig:contribution_stats}, the mean contribution increases substantially after the early stage, becomes largest around roughly 40k steps, and then gradually decreases toward the end of training.
The standard deviation exhibits a similar pattern, peaking in the middle stage and shrinking thereafter.
Since our contribution measure is defined as the reduction in distance to the final parameters, this suggests that updates in the middle stage play the largest role in moving the model toward its final state, while both early-stage and late-stage updates are less effective in this sense.

The opponent share in \cref{fig:opponent_share} provides a complementary perspective.
Opponents remain relatively rare through the early and middle stages, staying around 0--1\% for most of the trajectory before about 70k steps.
After that point, however, their share rises markedly, reaching roughly 6\% around 90k steps and around 8--10\% near the end of training.
This indicates that the late-stage decline in mean contribution is not explained solely by smaller positive contributions: increasingly many examples induce updates that are weakly aligned with, or even opposed to, the direction toward the final parameters.

Since the learning rate affects the scale of parameter updates, one might intuitively expect larger learning rates to be associated with larger contributions.
However, \cref{fig:contribution_stats} shows no simple one-to-one relationship between learning rate and contribution magnitude.
In particular, the rise in opponent share in \cref{fig:opponent_share} occurs while the learning rate is already decaying smoothly.
This is consistent with the definition in \cref{eq:cont_batch_closed}: contribution depends not only on the norm of the update but also on its alignment with the direction toward the final parameters. 
Thus, the later stages of training are characterized not only by smaller updates, but also by a growing fraction of opponent examples whose updates are less aligned with the eventual pretrained model.

\subsection{Relationship Between Text Difficulty and Contribution}
\label{sec:contribution_difficulty}

\begin{figure}[t]
\centering
\includegraphics[width=0.8\linewidth]{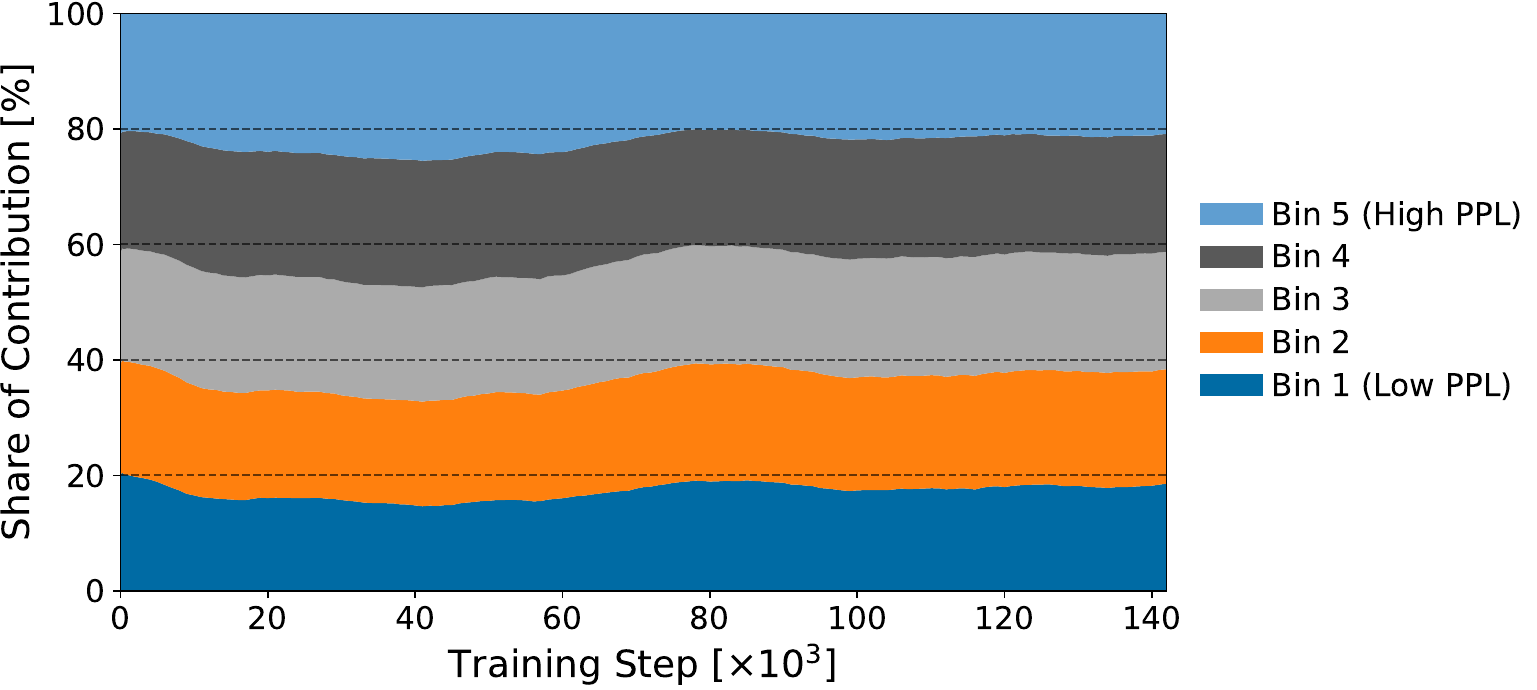}
\caption{
Relationship between text PPL and contribution for Pythia-1.4B-Deduped.
Each curve shows the share of normalized contribution attributed to an equal-sized PPL bin, with a moving-average window of 10 checkpoint intervals.
}
\label{fig:contribution_vs_ppl}
\end{figure}

Next, we focus on the relationship between training example difficulty and its contribution factor.
As a measure for a text's relative difficulty for a given model, we compute perplexity (PPL) using the final checkpoint of Pythia-12B-Deduped.
To place contributions on a positive scale, we apply min--max normalization within each checkpoint interval.
We then divide the 1,000 sampled examples in each checkpoint interval from \cref{sec:contribution_trajectory} into five equally sized bins according to PPL and compute the mean normalized contribution in each bin.

\Cref{fig:contribution_vs_ppl} shows that higher-PPL examples account for a larger share of normalized contribution during the middle stage of pretraining, while contribution is more evenly distributed across PPL bins during the early and late stages.
In particular, the share of the highest-PPL bin increases from approximately 20\% to 25\% in the middle stage, whereas that of the lowest-PPL bin decreases from approximately 20\% to 15\%.
Since all bins contain the same number of examples, this five-percentage-point shift represents a non-negligible redistribution of relative contribution toward higher-PPL data.

\subsection{Relationship Between Text Domain and Contribution}
\label{sec:contribution_domain}

\begin{figure}[t]
  \centering
  \captionsetup[subfigure]{justification=centering, skip=2pt}

  \begin{subfigure}{\linewidth}
    \centering
    \includegraphics[width=\linewidth]{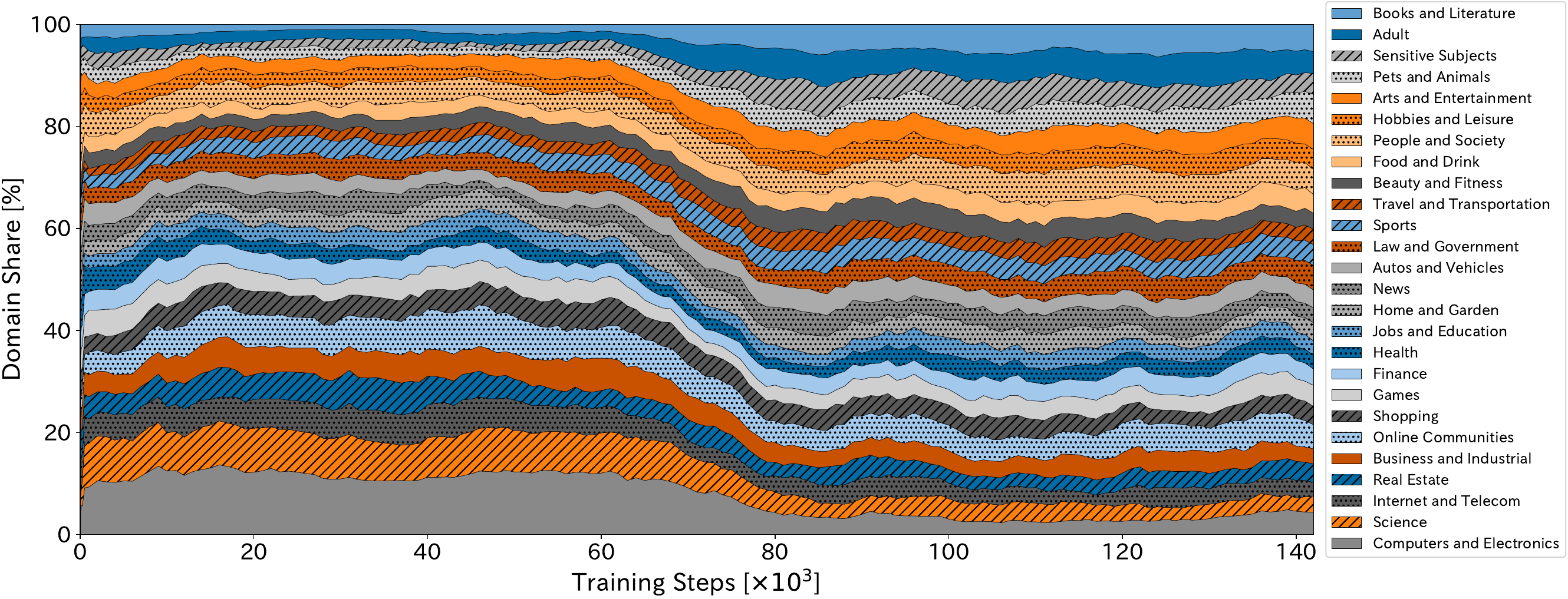}
    \caption{Proportion of domains among examples in the bottom 5\% of contributions.}
    \label{fig:bottom_contribution_count_by_domain}
  \end{subfigure}

  \vspace{0.5em}

  \begin{subfigure}{\linewidth}
    \centering
    \includegraphics[width=\linewidth]{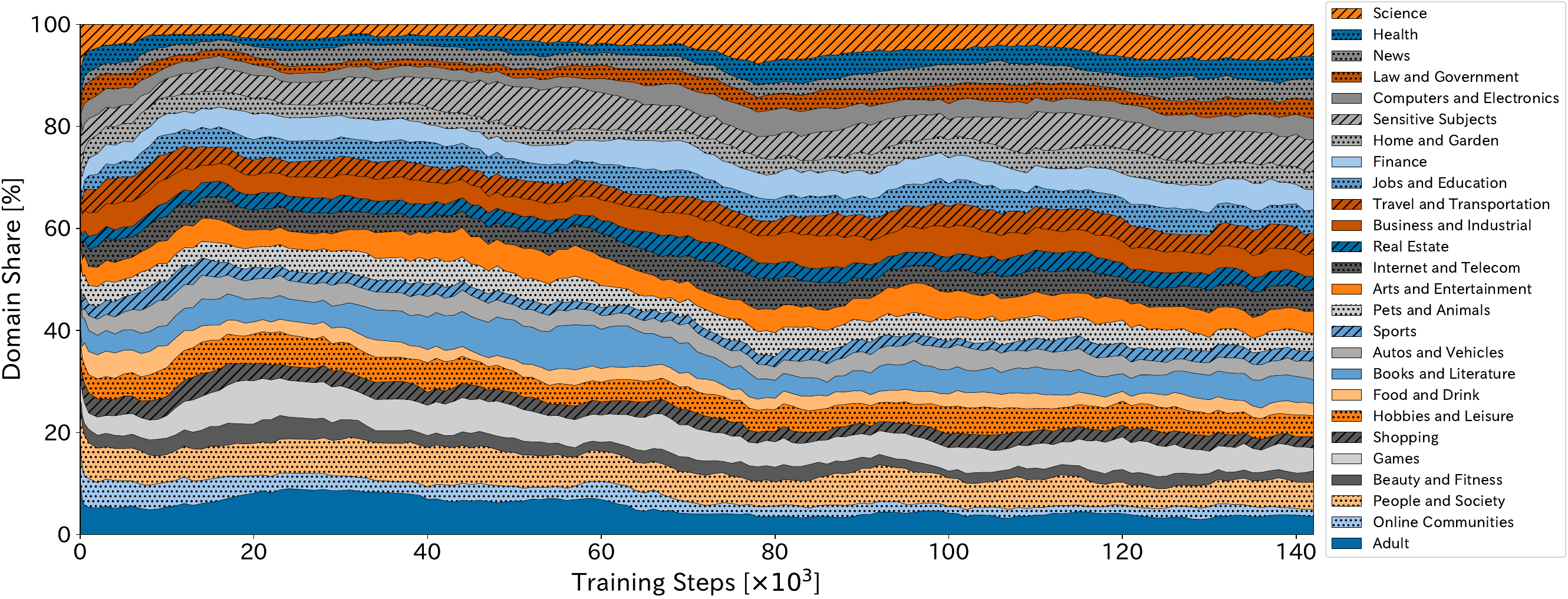}
    \caption{Proportion of domains among examples in the top 5\% of contributions.}
    \label{fig:top_contribution_count_by_domain}
  \end{subfigure}

  \caption{Domain composition of examples with extreme contribution values on Pythia-1.4B-Deduped. All curves are moving averages with window size 10.}
  \label{fig:contribution_count_by_domain}
\end{figure}

To obtain a more fine-grained view of which kinds of texts contribute to the final model, we next analyze the relationship between a training example's domain and its contribution.
To classify each example's domain, we use the NeMo Curator Domain Classifier,\footnote{\url{https://huggingface.co/nvidia/domain-classifier}} which assigns one of 26 domains, such as \textit{Finance} and \textit{Food and Drink}, based on a text's semantic content.
Rather than comparing average contributions across all examples, we focus on examples with extreme contribution values.
Specifically, for each checkpoint interval, we examine the domain composition of the bottom 5\% and top 5\% of examples ranked by contribution.
To avoid bias due to imbalanced domain frequencies, we randomly sample 100 examples from each domain in each checkpoint interval, resulting in 2,600 examples per interval for analysis.\footnote{Checkpoint intervals in which any domain has fewer than 100 examples are excluded from this analysis.}

\cref{fig:bottom_contribution_count_by_domain} shows that low-contribution examples are concentrated in a few domains during the first half of training, with STEM-related domains such as \textit{Computers and Electronics} and \textit{Science} accounting for a large proportion.
Under our contribution measure, this indicates that examples from these domains are relatively less aligned with progress toward the final parameters during the first half of training.
In the later stages, however, the shares of low-contribution STEM domains decrease, while domains such as \textit{Books and Literature} become more common among the low contributors.

\Cref{fig:top_contribution_count_by_domain} shows a complementary pattern.
Although STEM domains are not necessarily dominant among high-contribution examples early in training, their share tends to increase in later stages.
Together with the bottom-5\% results, this indicates that STEM-related data become increasingly aligned with progress toward the final parameters as pretraining proceeds.
Representative text excerpts from interval-level top and bottom contributors are provided in \cref{appendix:qualitative_examples}, giving a concrete view of the examples underlying these domain-level patterns.

Overall, these results reveal a systematic shift in which textual domains yield high contribution across pretraining.
In particular, literature-related data are relatively more prominent early in training, whereas STEM-related data become increasingly prominent in later stages.

\section{Analysis}
\label{sec:analysis}

We next evaluate the reliability and scope of our contribution analysis.
We validate the checkpoint-based approximation (\cref{sec:approximation_validation}), examine sensitivity to the reference endpoint (\cref{sec:reference_sensitivity}), assess robustness across model configurations (\cref{sec:cross_model_consistency}), and compare our measure with task-specific influence (\cref{sec:task_specific_comparison}).
We then discuss implications for pretraining strategies (\cref{sec:implications}).

\subsection{Validation of the Checkpoint-Based Approximation}
\label{sec:approximation_validation}

\begin{table}[t]
\centering
\small
\begin{tabular}{@{}lccc@{}}
\toprule
& \multicolumn{3}{c}{Checkpoint interval} \\
\cmidrule(l){2-4}
Metric
& Early: $1000\!\rightarrow\!2000$
& Mid: $10000\!\rightarrow\!11000$
& Late: $100000\!\rightarrow\!101000$ \\
\midrule
Pearson $r$  & $0.622^{*}$ & $0.808^{*}$ & $0.950^{*}$ \\
Spearman $r$ & $0.592^{*}$ & $0.811^{*}$ & $0.947^{*}$ \\
\bottomrule
\end{tabular}
\caption{
Correlation between exact example-level contribution and the checkpoint-based approximation on Pythia-70M-Deduped.
Superscript $^{*}$ indicates a correlation significantly different from zero under a two-sided test ($p<0.05$).
}
\label{tab:checkpoint_approx_validation}
\end{table}

Exact computation of example-level contribution requires access to the model parameters at every training step, whereas public model releases typically provide only sparse checkpoints.
We therefore evaluate how well the checkpoint-based approximation in \cref{eq:cont_approx_interval} matches the exact example-level contribution in \cref{eq:cont_sample_def} by retraining selected checkpoint intervals while storing all intermediate parameter states.
Because this procedure is computationally expensive, we perform this validation only on Pythia-70M-Deduped, considering three checkpoint intervals: 1k$\rightarrow$2k (early), 10k$\rightarrow$11k (middle), and 100k$\rightarrow$101k (late).
At each step in each interval, we sample 100 examples from the actual training data processed at that step and compute both the exact and approximate contributions for the same examples.
For each interval, we report the Pearson and Spearman correlations between the two quantities.

\cref{tab:checkpoint_approx_validation} shows positive correlations between the checkpoint-based approximation and the exact contribution in all three evaluated intervals, with substantially stronger agreement later in training.
This trend is intuitive: parameter changes between adjacent checkpoints are larger early in training and become smaller later, making the checkpoint-level approximation increasingly accurate.
Even in the early interval, the approximation shows moderate agreement with the exact contribution, with Pearson and Spearman correlations around 0.6.
For the sampled middle and late intervals, both correlations exceed 0.8, and they are above 0.94 in the late interval, indicating strong agreement in both contribution magnitudes and rankings.
Taken together, these results show that the checkpoint-based approximation tracks the exact example-level contribution increasingly well across the three evaluated stages of training.

\subsection{Sensitivity to the Reference Endpoint}
\label{sec:reference_sensitivity}

Our contribution score is defined relative to the final parameters of a particular pretraining run.
We therefore examine how the score changes when alternative checkpoints are used as the reference.
For Pythia-1.4B-Deduped, we use reference checkpoints at steps 30k, 70k, 120k, and 140k and compare the resulting contribution scores with the original scores defined using the final checkpoint at step 143k.
For each of five representative training intervals, i.e., 10k$\rightarrow$11k, 40k$\rightarrow$41k, 80k$\rightarrow$81k, 115k$\rightarrow$116k, and 130k$\rightarrow$131k, we use the same 1,000 examples sampled for the contribution analysis in \cref{sec:contribution_trajectory}.
For each alternative reference, we compute Pearson and Spearman correlations with the final-reference scores, as well as the overlap between their top and bottom 5\% sets, and average each metric across the five intervals.
The full results are reported in Appendix~\ref{appendix:reference_sensitivity}.

Using the near-final 140k checkpoint as the reference leaves the contribution rankings almost unchanged, yielding a mean Spearman correlation of 0.997 and a mean top-5\% overlap of 0.972 with the original scores.
By contrast, substantially earlier reference checkpoints produce markedly different rankings.
These results confirm that the contribution score is endpoint-dependent, while also showing that it is locally stable to the precise choice of a near-final reference checkpoint.

\subsection{Consistency of Contribution Dynamics across Model Configurations}
\label{sec:cross_model_consistency}

We next examine whether the contribution dynamics identified in \cref{sec:experiments} are consistent across model scales, weight initializations, and data orderings.

\paragraph{Model Sizes.}
Across the six Pythia-Deduped model sizes, the contribution-distribution, text-difficulty, and domain analyses exhibit broadly similar stage-dependent dynamics, while also showing systematic scale-dependent differences (see Appendix~\ref{appendix:model_size}).
In particular, several transitions that occur during the middle stage for larger models appear later for smaller models, suggesting that the timing of training-phase transitions may depend on model scale.
We also observe that the magnitude of the contribution dynamics tends to be smaller at larger scales.
Thus, model size appears to affect when and how strongly contribution patterns change across pretraining, while preserving much of their broader temporal structure.

\paragraph{Weight Initialization and Data Ordering.}
The qualitative domain-level contribution dynamics are also broadly consistent across PolyPythia runs with different random factors (see Appendix~\ref{appendix:random_initialization}).
Across the standard 160M and 410M runs, similar temporal patterns are preserved across seeds, except for the anomalous 410M seed-3 run, which \citet{wal2025polypythias} identify as an outlier exhibiting loss spikes and degraded performance.
The decoupled PolyPythia-160M variants further show highly consistent dynamics when only weight initialization is varied, while changing data ordering introduces somewhat greater variation.
Overall, these results suggest that the stage-dependent contribution patterns are robust to ordinary seed variation, although data ordering may have a somewhat larger effect than weight initialization.

\subsection{Comparison with Task-Specific Influence}
\label{sec:task_specific_comparison}

To examine how our contribution measure, which is agnostic to the choice of downstream task or validation set, differs from domain-specific attribution, we compare it with a TracIn-style influence score~\citep{pruthi2020tracin} computed with respect to domain-wise language-modeling loss.
Specifically, we measure the loss on domain-specific held-out validation sets from The Pile, obtained by filtering out examples that occur in the Pythia pretraining stream.
Using the same domain classifier as in \cref{sec:contribution_domain}, we then sample 1,000 high-confidence examples for each of the four validation domains: \textit{Books and Literature}, \textit{Science}, \textit{Computers and Electronics}, and \textit{Hobbies and Leisure}.
For each checkpoint interval, we use the same domain-balanced training examples as in \cref{sec:contribution_domain} and compute the gradient alignment between each training example's loss and the language-modeling loss on each domain-specific validation set.
We then examine the domain composition of the top and bottom 5\% of examples under the resulting TracIn-style scores, analogously to \cref{sec:contribution_domain}.
The full experimental results are provided in Appendix~\ref{appendix:task_specific_comparison}.

The resulting attribution patterns depend substantially on the selected validation domain.
Training examples from the selected validation domain are strongly overrepresented among the highest-scoring examples early in training, particularly when \textit{Science} is used as the validation target, although this enrichment becomes more diffuse in later stages.
When non-STEM domains such as \textit{Books and Literature} or \textit{Hobbies and Leisure} are used as validation targets, STEM-related training examples are overrepresented among the lowest-scoring examples early in training.
Thus, the domain-specific TracIn analysis does not consistently recover the Literature-to-STEM crossover observed with our contribution measure.

\subsection{Implications for Pretraining Strategies}
\label{sec:implications}

We conclude our analysis by discussing how the observed contribution dynamics inform our understanding of pretraining and the design of pretraining data strategies.

\paragraph{Pretraining Phases.}
For Pythia, \citet{wal2025polypythias} report that all models exhibit an \textit{initial learning phase} between 1k--10k steps in which baseline linguistic capabilities are acquired, as well as a \textit{critical learning phase} at 10k--100k steps during which most downstream performance gains occur.
Our contribution dynamics shed additional light on these phases: higher-PPL texts and STEM-related domains tend to contribute less during the initial learning phase but more strongly during the critical learning phase.

\paragraph{Stage-Aware Data Composition.}
A common view in curriculum learning~\citep{bengio2009curriculum,wang2022survey} is that training should proceed from easy to difficult examples.
Our results do not fully support this simple progression: difficult texts contribute more strongly to moving the model toward the final parameters primarily during the middle stage, rather than becoming progressively more important toward the end of training.
In practice, recent LLM pretraining recipes adopt a more specific stage-aware heuristic by increasing the allocation of STEM-related data, such as math and code, in later stages of training~\citep{martins2024eurollmm,bakouch2025smollm3}.
Our contribution dynamics lend empirical support to this heuristic from a trajectory-level perspective, as STEM-related domains become increasingly prominent among high-contribution examples later in pretraining.
More broadly, they motivate pretraining strategies that adapt data composition across training stages rather than treating a single data mixture as equally suitable throughout training.

\paragraph{Beyond Task-Specific Evaluation.}
Task-specific evaluations provide valuable signals for assessing pretraining data, but the resulting conclusions can depend on the choice of evaluation target.
For instance, \citet{jiao2025datelm} show that the evaluation of data attribution methods is sensitive to task-specific evaluation design.
Similarly, \citet{rosa-etal-2025-impact} report heterogeneous downstream effects across different types of literary data, while our contribution analysis identifies literature-related data as prominent contributors during early pretraining.
We hope that our task-agnostic contribution measure can reveal stage-dependent roles of training data beyond fixed downstream evaluations and help guide more nuanced pretraining strategies.

\section{Conclusion}
\label{sec:conclusion}

In this work, we proposed a task-agnostic measure of training data influence by defining an example's contribution as the reduction in distance to the final parameters of a given pretraining run.
Across model configurations, we found systematic contribution dynamics throughout pretraining, including earlier transitions for larger models, greater relative contribution from difficult texts during the middle stage, and a crossover from literature-related to STEM-related domains toward later training.
Together, these results provide a trajectory-level view of how the training data most strongly aligned with progress toward the final parameters change across pretraining, complementing attribution analyses defined with respect to particular downstream tasks or validation sets.

\section*{Acknowledgments}
This study is partially supported by JST BOOST JPMJBY24A6. 
Max Müller-Eberstein is supported by the Carlsberg Foundation, grant CF-25-0624.
Computational resources were provided in part by ``mdx: a platform for building data-empowered society.''

\bibliography{colm2026_conference}

@InProceedings{pmlr-v70-koh17a,
  title = 	 {Understanding Black-box Predictions via Influence Functions},
  author =       {Pang Wei Koh and Percy Liang},
  booktitle = 	 {Proceedings of the 34th International Conference on Machine Learning},
  pages = 	 {1885--1894},
  year = 	 {2017},
  editor = 	 {Precup, Doina and Teh, Yee Whye},
  volume = 	 {70},
  series = 	 {Proceedings of Machine Learning Research},
  month = 	 {06--11 Aug},
  publisher =    {PMLR},
  url = 	 {https://proceedings.mlr.press/v70/koh17a.html}
}

@inproceedings{pruthi2020tracin,
 author = {Pruthi, Garima and Liu, Frederick and Kale, Satyen and Sundararajan, Mukund},
 booktitle = {Advances in Neural Information Processing Systems},
 editor = {H. Larochelle and M. Ranzato and R. Hadsell and M.F. Balcan and H. Lin},
 pages = {19920--19930},
 publisher = {Curran Associates, Inc.},
 title = {Estimating Training Data Influence by Tracing Gradient Descent},
 url = {https://proceedings.neurips.cc/paper_files/paper/2020/file/e6385d39ec9394f2f3a354d9d2b88eec-Paper.pdf},
 volume = {33},
 year = {2020}
}

@article{hammoudeh2024survey,
author={Hammoudeh, Zayd
and Lowd, Daniel},
title={Training data influence analysis and estimation: a survey},
journal={Machine Learning},
year={2024},
month={May},
day={01},
volume={113},
number={5},
pages={2351-2403},
issn={1573-0565},
doi={10.1007/s10994-023-06495-7},
url={https://doi.org/10.1007/s10994-023-06495-7}
}

@inproceedings{
chang2025scalable,
title={Scalable Influence and Fact Tracing for Large Language Model Pretraining},
author={Tyler A. Chang and Dheeraj Rajagopal and Tolga Bolukbasi and Lucas Dixon and Ian Tenney},
booktitle={The Thirteenth International Conference on Learning Representations},
year={2025},
url={https://openreview.net/forum?id=gLa96FlWwn}
}

@inproceedings{
wang2025data,
title={Data Shapley in One Training Run},
author={Jiachen T. Wang and Prateek Mittal and Dawn Song and Ruoxi Jia},
booktitle={The Thirteenth International Conference on Learning Representations},
year={2025},
url={https://openreview.net/forum?id=HD6bWcj87Y}
}

@InProceedings{pmlr-v202-biderman23a,
  title = 	 {Pythia: A Suite for Analyzing Large Language Models Across Training and Scaling},
  author =       {Biderman, Stella and Schoelkopf, Hailey and Anthony, Quentin Gregory and Bradley, Herbie and O'Brien, Kyle and Hallahan, Eric and Khan, Mohammad Aflah and Purohit, Shivanshu and Prashanth, Usvsn Sai and Raff, Edward and Skowron, Aviya and Sutawika, Lintang and Van Der Wal, Oskar},
  booktitle = 	 {Proceedings of the 40th International Conference on Machine Learning},
  pages = 	 {2397--2430},
  year = 	 {2023},
  editor = 	 {Krause, Andreas and Brunskill, Emma and Cho, Kyunghyun and Engelhardt, Barbara and Sabato, Sivan and Scarlett, Jonathan},
  volume = 	 {202},
  series = 	 {Proceedings of Machine Learning Research},
  month = 	 {23--29 Jul},
  publisher =    {PMLR},
  url = 	 {https://proceedings.mlr.press/v202/biderman23a.html}
}

@misc{gao2020pile800gbdatasetdiverse,
      title={The Pile: An 800GB Dataset of Diverse Text for Language Modeling}, 
      author={Leo Gao and Stella Biderman and Sid Black and Laurence Golding and Travis Hoppe and Charles Foster and Jason Phang and Horace He and Anish Thite and Noa Nabeshima and Shawn Presser and Connor Leahy},
      year={2020},
      eprint={2101.00027},
      archivePrefix={arXiv},
      primaryClass={cs.CL},
      url={https://arxiv.org/abs/2101.00027}, 
}

@InProceedings{pmlr-v97-ghorbani19c,
  title = 	 {Data Shapley: Equitable Valuation of Data for Machine Learning},
  author =       {Ghorbani, Amirata and Zou, James},
  booktitle = 	 {Proceedings of the 36th International Conference on Machine Learning},
  pages = 	 {2242--2251},
  year = 	 {2019},
  editor = 	 {Chaudhuri, Kamalika and Salakhutdinov, Ruslan},
  volume = 	 {97},
  series = 	 {Proceedings of Machine Learning Research},
  month = 	 {09--15 Jun},
  publisher =    {PMLR},
  url = 	 {https://proceedings.mlr.press/v97/ghorbani19c.html}
}

@InProceedings{pmlr-v202-park23c,
  title = 	 {{TRAK}: Attributing Model Behavior at Scale},
  author =       {Park, Sung Min and Georgiev, Kristian and Ilyas, Andrew and Leclerc, Guillaume and Madry, Aleksander},
  booktitle = 	 {Proceedings of the 40th International Conference on Machine Learning},
  pages = 	 {27074--27113},
  year = 	 {2023},
  editor = 	 {Krause, Andreas and Brunskill, Emma and Cho, Kyunghyun and Engelhardt, Barbara and Sabato, Sivan and Scarlett, Jonathan},
  volume = 	 {202},
  series = 	 {Proceedings of Machine Learning Research},
  month = 	 {23--29 Jul},
  publisher =    {PMLR},
  url = 	 {https://proceedings.mlr.press/v202/park23c.html}
}

@inproceedings{bengio2009curriculum,
author = {Bengio, Yoshua and Louradour, J\'{e}r\^{o}me and Collobert, Ronan and Weston, Jason},
title = {Curriculum learning},
year = {2009},
isbn = {9781605585161},
publisher = {Association for Computing Machinery},
address = {New York, NY, USA},
url = {https://doi.org/10.1145/1553374.1553380},
doi = {10.1145/1553374.1553380},
booktitle = {Proceedings of the 26th Annual International Conference on Machine Learning},
pages = {41–48},
numpages = {8},
location = {Montreal, Quebec, Canada},
series = {ICML '09}
}

@article{wang2022survey,
author={Wang, Xin and Chen, Yudong and Zhu, Wenwu},
journal={ IEEE Transactions on Pattern Analysis \& Machine Intelligence },
title={{ A Survey on Curriculum Learning }},
year={2022},
volume={44},
number={09},
ISSN={1939-3539},
pages={4555-4576},
doi={10.1109/TPAMI.2021.3069908},
url = {https://doi.ieeecomputersociety.org/10.1109/TPAMI.2021.3069908},
publisher={IEEE Computer Society},
address={Los Alamitos, CA, USA},
month=sep}

@inproceedings{
jiao2025datelm,
title={{DATE}-{LM}: Benchmarking Data Attribution Evaluation for Large Language Models},
author={Cathy Jiao and Yijun Pan and Emily Xiao and Daisy Sheng and Niket Jain and Hanzhang Zhao and Ishita Dasgupta and Jiaqi W. Ma and Chenyan Xiong},
booktitle={The Thirty-ninth Annual Conference on Neural Information Processing Systems Datasets and Benchmarks Track},
year={2025},
url={https://openreview.net/forum?id=e2cD5xuHix}
}

@inproceedings{
choe2025what,
title={What is Your Data Worth to {GPT}? {LLM}-Scale Data Valuation with Influence Functions},
author={Sang Keun Choe and Hwijeen Ahn and Juhan Bae and Kewen Zhao and Youngseog Chung and Adithya Pratapa and Willie Neiswanger and Emma Strubell and Teruko Mitamura and Jeff Schneider and Eduard Hovy and Roger Baker Grosse and Eric P. Xing},
booktitle={The Thirty-ninth Annual Conference on Neural Information Processing Systems},
year={2025},
url={https://openreview.net/forum?id=zPKeJAEo27}
}

@inproceedings{kingma2015adam,
  author    = {Kingma, Diederik P. and Ba, Jimmy},
  title     = {Adam: A Method for Stochastic Optimization},
  booktitle = {International Conference on Learning Representations (ICLR)},
  year      = {2015}
}

@inproceedings{
wal2025polypythias,
title={PolyPythias: Stability and Outliers across Fifty Language Model Pre-Training Runs},
author={Oskar van der Wal and Pietro Lesci and Max M{\"u}ller-Eberstein and Naomi Saphra and Hailey Schoelkopf and Willem Zuidema and Stella Biderman},
booktitle={The Thirteenth International Conference on Learning Representations},
year={2025},
url={https://openreview.net/forum?id=bmrYu2Ekdz}
}

@inproceedings{
cazenavette2022distillation,
title={Dataset Distillation by Matching Training Trajectories},
author={George Cazenavette and Tongzhou Wang and Antonio Torralba and Alexei A. Efros and Jun-Yan Zhu},
booktitle={Proceedings of the IEEE/CVF Conference on Computer Vision and Pattern Recognition},
year={2022}
}

@InProceedings{pmlr-v202-cui23e,
  title = 	 {Scaling Up Dataset Distillation to {I}mage{N}et-1{K} with Constant Memory},
  author =       {Cui, Justin and Wang, Ruochen and Si, Si and Hsieh, Cho-Jui},
  booktitle = 	 {Proceedings of the 40th International Conference on Machine Learning},
  pages = 	 {6565--6590},
  year = 	 {2023},
  editor = 	 {Krause, Andreas and Brunskill, Emma and Cho, Kyunghyun and Engelhardt, Barbara and Sabato, Sivan and Scarlett, Jonathan},
  volume = 	 {202},
  series = 	 {Proceedings of Machine Learning Research},
  month = 	 {23--29 Jul},
  publisher =    {PMLR},
  url = 	 {https://proceedings.mlr.press/v202/cui23e.html}
}

@InProceedings{liu2024dataset,
author="Liu, Dai
and Gu, Jindong
and Cao, Hu
and Trinitis, Carsten
and Schulz, Martin",
editor="Leonardis, Ale{\v{s}}
and Ricci, Elisa
and Roth, Stefan
and Russakovsky, Olga
and Sattler, Torsten
and Varol, G{\"u}l",
title="Dataset Distillation by Automatic Training Trajectories",
booktitle="Computer Vision -- ECCV 2024",
year="2025",
publisher="Springer Nature Switzerland",
address="Cham",
pages="334--351",
isbn="978-3-031-73021-4"
}

@InProceedings{pmlr-v162-ilyas22a,
  title = 	 {Datamodels: Understanding Predictions with Data and Data with Predictions},
  author =       {Ilyas, Andrew and Park, Sung Min and Engstrom, Logan and Leclerc, Guillaume and Madry, Aleksander},
  booktitle = 	 {Proceedings of the 39th International Conference on Machine Learning},
  pages = 	 {9525--9587},
  year = 	 {2022},
  editor = 	 {Chaudhuri, Kamalika and Jegelka, Stefanie and Song, Le and Szepesvari, Csaba and Niu, Gang and Sabato, Sivan},
  volume = 	 {162},
  series = 	 {Proceedings of Machine Learning Research},
  month = 	 {17--23 Jul},
  publisher =    {PMLR},
  url = 	 {https://proceedings.mlr.press/v162/ilyas22a.html}
}

@article{pan2025detecting,
  title = {Detecting and Filtering Unsafe Training Data via Data Attribution with Denoised Representation},
  author = {Pan, Yijun and Shi, Taiwei and Zhao, Jieyu and Ma, Jiaqi W.},
  journal = {arXiv preprint arXiv:2502.11411},
  year = {2025},
  doi = {10.48550/arXiv.2502.11411},
  url = {https://arxiv.org/abs/2502.11411}
}

@inproceedings{
yu2024mates,
title={{MATES}: Model-Aware Data Selection for Efficient Pretraining with Data Influence Models},
author={Zichun Yu and Spandan Das and Chenyan Xiong},
booktitle={The Thirty-eighth Annual Conference on Neural Information Processing Systems},
year={2024},
url={https://openreview.net/forum?id=6gzPSMUAz2}
}

@inproceedings{
xia2024less,
title={{LESS}: Selecting Influential Data for Targeted Instruction Tuning},
author={Mengzhou Xia and Sadhika Malladi and Suchin Gururangan and Sanjeev Arora and Danqi Chen},
booktitle={Forty-first International Conference on Machine Learning},
year={2024},
url={https://openreview.net/forum?id=PG5fV50maR}
}

@inproceedings{akyurek-etal-2022-towards,
    title = "Towards Tracing Knowledge in Language Models Back to the Training Data",
    author = "Akyurek, Ekin  and
      Bolukbasi, Tolga  and
      Liu, Frederick  and
      Xiong, Binbin  and
      Tenney, Ian  and
      Andreas, Jacob  and
      Guu, Kelvin",
    editor = "Goldberg, Yoav  and
      Kozareva, Zornitsa  and
      Zhang, Yue",
    booktitle = "Findings of the Association for Computational Linguistics: EMNLP 2022",
    month = dec,
    year = "2022",
    address = "Abu Dhabi, United Arab Emirates",
    publisher = "Association for Computational Linguistics",
    url = "https://aclanthology.org/2022.findings-emnlp.180/",
    doi = "10.18653/v1/2022.findings-emnlp.180",
    pages = "2429--2446"
}

@inproceedings{rosa-etal-2025-impact,
    title = "The Impact of Copyrighted Material on Large Language Models: {A} {Norwegian} Perspective",
    author = "de la Rosa, Javier  and
      Mikhailov, Vladislav  and
      Zhang, Lemei  and
      Wetjen, Freddy  and
      Samuel, David  and
      Liu, Peng  and
      Braaten, Rolv-Arild  and
      M{\ae}hlum, Petter  and
      Birkenes, Magnus Breder  and
      Kutuzov, Andrey  and
      Enstad, Tita  and
      Farseth{\r{a}}s, Hans Christian  and
      Brygfjeld, Svein Arne  and
      Gulla, Jon Atle  and
      Oepen, Stephan  and
      Velldal, Erik  and
      {\O}stgulen, Wilfred  and
      {\O}vrelid, Lilja  and
      Myhre, Aslak Sira",
    editor = "Johansson, Richard  and
      Stymne, Sara",
    booktitle = "Proceedings of the Joint 25th Nordic Conference on Computational Linguistics and 11th Baltic Conference on Human Language Technologies (NoDaLiDa/Baltic-HLT 2025)",
    month = mar,
    year = "2025",
    address = "Tallinn, Estonia",
    publisher = "University of Tartu Library",
    url = "https://aclanthology.org/2025.nodalida-1.59/",
    pages = "544--560",
    ISBN = "978-9908-53-109-0"
}

@misc{bakouch2025smollm3,
  title={{SmolLM3: smol, multilingual, long-context reasoner}},
  author={Bakouch, Elie and Ben Allal, Loubna and Lozhkov, Anton and Tazi, Nouamane and Tunstall, Lewis and Patiño, Carlos Miguel and Beeching, Edward and Roucher, Aymeric and Reedi, Aksel Joonas and Gallouédec, Quentin and Rasul, Kashif and Habib, Nathan and Fourrier, Clémentine and Kydlicek, Hynek and Penedo, Guilherme and Larcher, Hugo and Morlon, Mathieu and Srivastav, Vaibhav and Lochner, Joshua and Nguyen, Xuan-Son and Raffel, Colin and von Werra, Leandro and Wolf, Thomas},
  year={2025},
  howpublished={\url{https://huggingface.co/blog/smollm3}}
}

@misc{martins2024eurollmm,
      title={EuroLLM: Multilingual Language Models for Europe}, 
      author={Pedro Henrique Martins and Patrick Fernandes and João Alves and Nuno M. Guerreiro and Ricardo Rei and Duarte M. Alves and José Pombal and Amin Farajian and Manuel Faysse and Mateusz Klimaszewski and Pierre Colombo and Barry Haddow and José G. C. de Souza and Alexandra Birch and André F. T. Martins},
      year={2024},
      eprint={2409.16235},
      archivePrefix={arXiv},
      primaryClass={cs.CL},
      url={https://arxiv.org/abs/2409.16235}, 
}
\bibliographystyle{colm2026_conference}

\appendix

\section{Limitations}
\label{appendix:limitations}

Our contribution measure does not require selecting a downstream task or validation set as the attribution target, but it remains defined relative to the final parameters of a particular pretraining run.
As shown in \cref{sec:reference_sensitivity}, substantially different reference checkpoints can yield different contribution rankings.
Moreover, squared parameter-space distance provides a scalable geometric notion of progress, but it is sensitive to parameterization and does not directly measure changes in model behavior.
Function-aware alternatives based on predictive divergence or information geometry could provide complementary notions of contribution, but are substantially more expensive at pretraining scale.

Our checkpoint-based approximation replaces step-level model states with periodically saved checkpoints.
We validate this approximation on three intervals of Pythia-70M-Deduped and observe stronger agreement later in training, but exact validation at larger model scales remains computationally expensive.
In addition, the example-level decomposition assumes SGD, whereas the models analyzed here are trained with adaptive optimizers.
Accounting exactly for optimizer states and their history would require a substantially more involved decomposition.

Although the qualitative contribution dynamics are broadly consistent across the Pythia and PolyPythia configurations we examine, all experiments remain within closely related model families and pretraining data.
Their generality to substantially different architectures, datasets, and training recipes therefore remains to be established.
Our analyses also characterize how contribution varies across training rather than directly establishing the causal effect of changing the data mixture.
Accordingly, the stage-dependent patterns observed here motivate data-composition and curriculum strategies, but their effects on downstream performance require direct intervention-based evaluation.
Finally, disentangling domain effects from correlated properties such as text difficulty remains an important direction for future work.

\section{Additional Results}
\label{appendix:additional_results}

\subsection{Qualitative Examples for Pretraining Domains}
\label{appendix:qualitative_examples}

The examples in \cref{tab:domain_qualitative_examples} qualitatively mirror the aggregate domain-level patterns in \cref{sec:contribution_domain}.
Early in training, low-contribution examples include highly technical or structurally complex texts, such as symbolic scientific reasoning and markup-heavy computer-related content, whereas the high-contribution example is conversational and closer to natural discourse.
Later in training, technical scientific and computer-related passages appear among the top contributors, while a narrative passage appears among the bottom contributors.
They provide concrete illustrations of the broader temporal shift observed in our domain-level analysis.

\begin{table}[t]
\centering
\footnotesize
\setlength{\tabcolsep}{4pt}
\renewcommand{\arraystretch}{1.12}
\begin{tabularx}{\linewidth}{@{} l l >{\raggedright\arraybackslash}X @{}}
\toprule
 & Interval & Domain / Text excerpt \\
\midrule

Bottom & 25k$\rightarrow$26k 
& \texttt{[Science]} \quad
\textit{\ldots
The axioms TKF10 and TKF11 allow us to infer that
$Dy(ClTm(y) \wedge T x\psi(\dot y/x)y) \wedge @y(ClTm(y) \to T x \psi(\dot y/x)y)$
must hold at $w$. Let $y_0$ witness the first conjunct, and specialize the second conjunct to it. We get that at $w$, it must be that
$M( T x\psi(\dot y_0/x)y \wedge T x \psi(\dot y_0/x)y )[w]$.
\ldots} \\

Bottom & 4k$\rightarrow$5k 
& \texttt{[Computers \& Electronics]} \quad
\textit{\ldots
\texttt{<a class="" href="../../../demos/logger-strategy/">Logger</a>}
\textbackslash n\hspace{0pt}
\texttt{</li>}
\textbackslash n\hspace{0pt}
\texttt{<li class="toctree-l3">}
\textbackslash n\hspace{0pt}
\texttt{<a class="" href="../../../demos/accesslog/">Accesslog</a>}
\textbackslash n\hspace{0pt}
\texttt{</li>}
\ldots} \\

Top & 5k$\rightarrow$6k 
& \texttt{[Hobbies \& Leisure]} \quad
\textit{\ldots
Kaminsky: Yeah, exactly.
\textbackslash n\textbackslash n
AJK: And they kept through, right, yeah, which is -- yeah, and again some of your -- I just, you know, see for the first time which is kind of exciting to me, to not have the prefabricated opinions on it at a time.
\ldots} \\

\midrule

Top & 107k$\rightarrow$108k
& \texttt{[Science]} \quad
\textit{\ldots
Final control parameters ($\sigma$ values) of input data sets and potential terms.
\textbackslash n\textbackslash n
In an additional reference series of RMC calculations the FNC method has been used again and in addition, partial pair correlation functions ($g_{ij}(r)$) from MD simulations have been
\ldots} \\

Top & 114k$\rightarrow$115k
& \texttt{[Computers \& Electronics]} \quad
\textit{\ldots
For example, if the patient object in [Figure 3](\#sensors-20-02264-f003)\{ref-type="fig"\} occupies $400 \times 400$ number of pixels in an omnidirectional view, then the patient in the foveal region of foveated view will still be rendered on $400 \times 400$ number of pixels
\ldots} \\

Bottom & 107k$\rightarrow$108k
& \texttt{[Books \& Literature]} \quad
\textit{\ldots
``I've the best news,'' Khiri said.
\textbackslash n\textbackslash n
Hal, who was still brooding about his seemingly unavoidable retirement, grumped something, which Carstares took as interest.
\textbackslash n\textbackslash n
``You'll be permitted to leave the hospital under my care any day now,'' she said.
\ldots} \\

\bottomrule
\end{tabularx}
\caption{Qualitative domain examples drawn from the interval-level top-10 and bottom-10 examples by approximate contribution. Top/Bottom indicates whether the example belongs to the top-10 or bottom-10 within the corresponding checkpoint interval. Text excerpts are lightly truncated and normalized for readability.}
\label{tab:domain_qualitative_examples}
\end{table}

\subsection{Full Results for Reference-Endpoint Sensitivity}
\label{appendix:reference_sensitivity}

\begin{table}[t]
\centering
\small
\begin{tabular}{@{}lrrrr@{}}
\toprule
Reference & Spearman & Pearson & Top-5\% overlap & Bottom-5\% overlap \\
\midrule
30k  & $-0.554$ & $-0.559$ & $0.064$ & $0.148$ \\
70k  & $-0.174$ & $-0.173$ & $0.316$ & $0.352$ \\
120k & $0.566$  & $0.573$  & $0.680$ & $0.656$ \\
140k & $0.997$  & $0.998$  & $0.972$ & $0.964$ \\
\bottomrule
\end{tabular}
\caption{
Full reference-sensitivity results on Pythia-1.4B-Deduped.
Each alternative-reference score is compared with the original score defined using the final checkpoint at step 143k.
All reported correlations and overlaps are means across the five checkpoint intervals.
}
\label{tab:reference_sensitivity}
\end{table}

We report the full results for the reference-sensitivity analysis summarized in \cref{sec:reference_sensitivity}.
We evaluate the checkpoint intervals 10k$\rightarrow$11k, 40k$\rightarrow$41k, 80k$\rightarrow$81k, 115k$\rightarrow$116k, and 130k$\rightarrow$131k, using the same 1,000 examples sampled for the contribution analysis in \cref{sec:contribution_trajectory}.
For each alternative reference checkpoint, we compute Pearson and Spearman correlations with the scores obtained using the final 143k checkpoint, as well as the overlap between their top and bottom 5\% sets.
\cref{tab:reference_sensitivity} reports the mean of each metric across the five intervals.

\subsection{Consistency across Model Sizes}
\label{appendix:model_size}

We extend the analyses in \cref{sec:contribution_trajectory,sec:contribution_difficulty,sec:contribution_domain} to all six Pythia-Deduped model sizes: 70M, 160M, 410M, 1.4B, 6.9B, and 12B.
Across scales, the broad temporal patterns of contribution are qualitatively consistent, although their timing and magnitude vary with model size.

\begin{figure*}[t]
  \centering
  \captionsetup[subfigure]{justification=centering, skip=2pt}

  \begin{subfigure}{\textwidth}
    \centering
    \includegraphics[width=\linewidth]{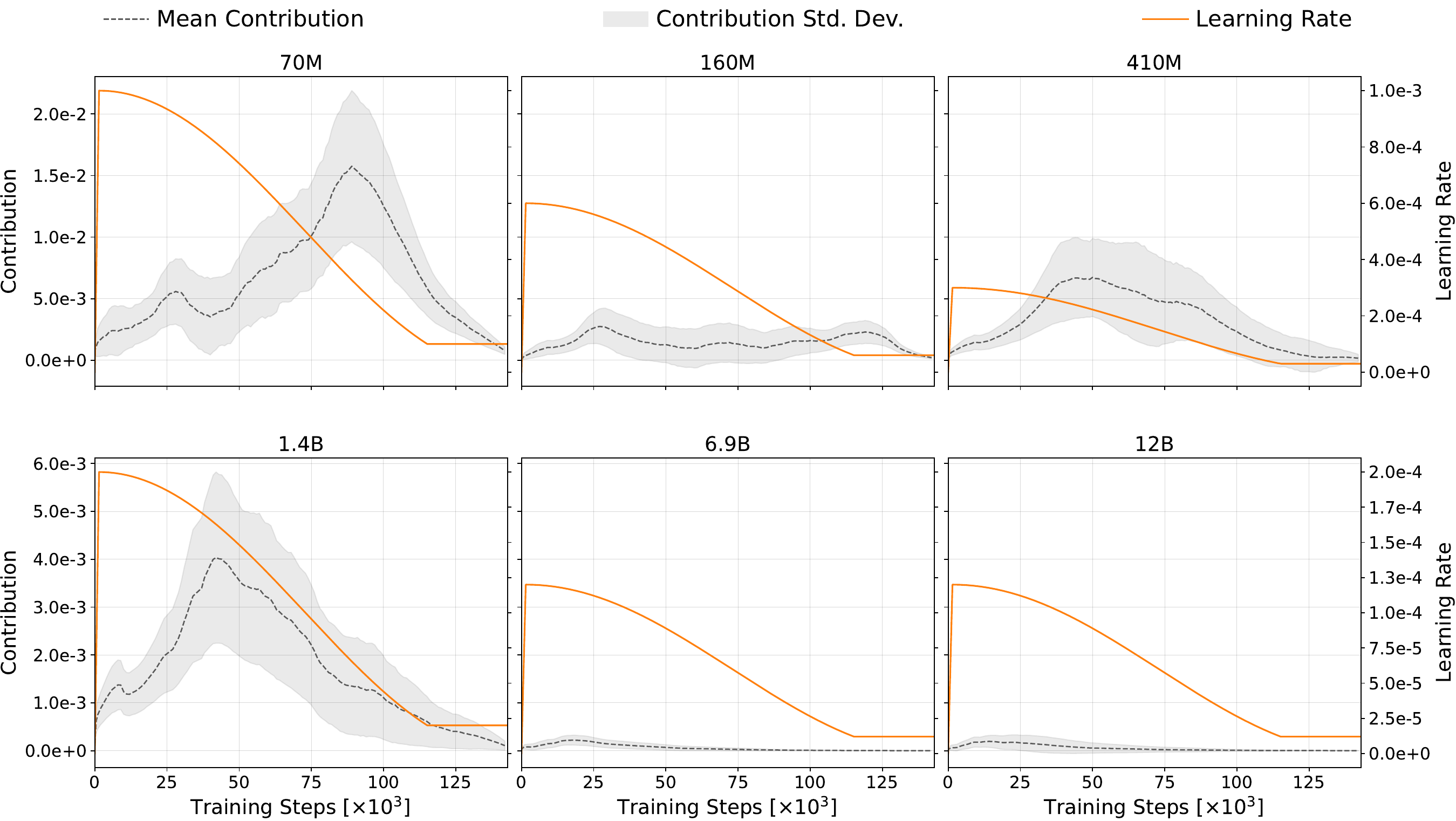}
    \caption{
    Contribution statistics over training.
    }
    \label{fig:contribution_stats_across_sizes}
  \end{subfigure}

  \vspace{0.5em}

  \begin{subfigure}{\textwidth}
    \centering
    \includegraphics[width=0.9\linewidth]{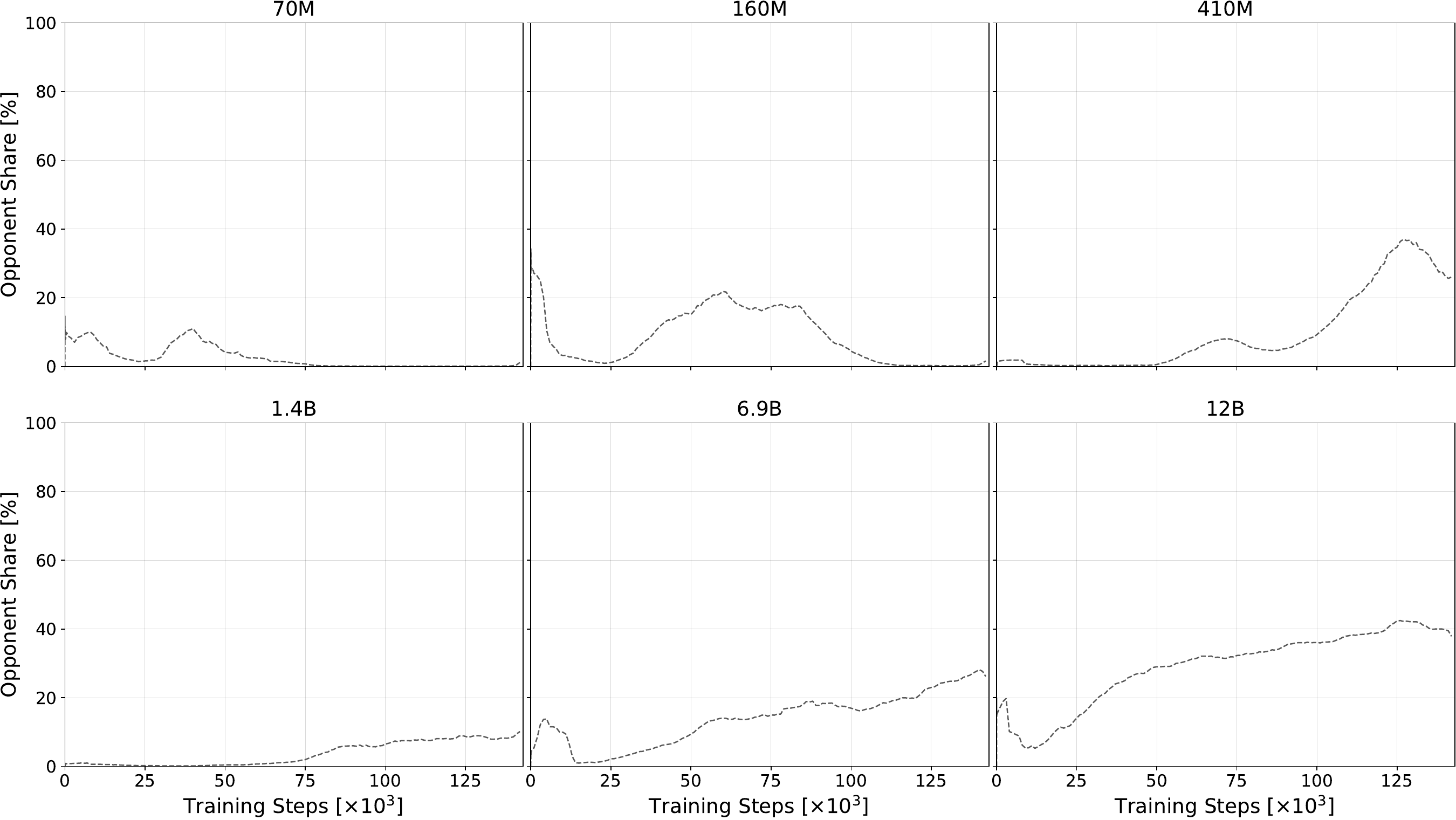}
    \caption{Share of opponent examples.}
    \label{fig:opponent_share_across_sizes}
  \end{subfigure}

  \caption{
  Training dynamics of the contribution distribution across the six Pythia-Deduped model sizes.
  Each panel corresponds to one model size.
  All curves are moving averages with window size 10.
  }
  \label{fig:contribution_dynamics_across_sizes}
\end{figure*}

\Cref{fig:contribution_dynamics_across_sizes} extends the contribution-distribution analysis in \cref{sec:contribution_trajectory} across model sizes.
Except for the 160M model, mean contribution is highest during the middle stage of pretraining, although the location of the peak varies somewhat across model sizes.
The 160M model instead exhibits a bimodal pattern, with higher contribution in both the early and late stages.
We also observe that the absolute magnitude of contribution tends to decrease with model size, which may partly reflect differences in the learning-rate schedules used across scales.
For models at 410M and above, the share of opponent examples consistently tends to increase toward the end of training, although its absolute level varies substantially across model sizes.
Overall, the broad temporal structure of the contribution distribution is shared across most scales, while its precise shape, magnitude, and opponent share remain model-dependent.

\begin{figure*}[t]
  \centering
  \includegraphics[width=\textwidth]{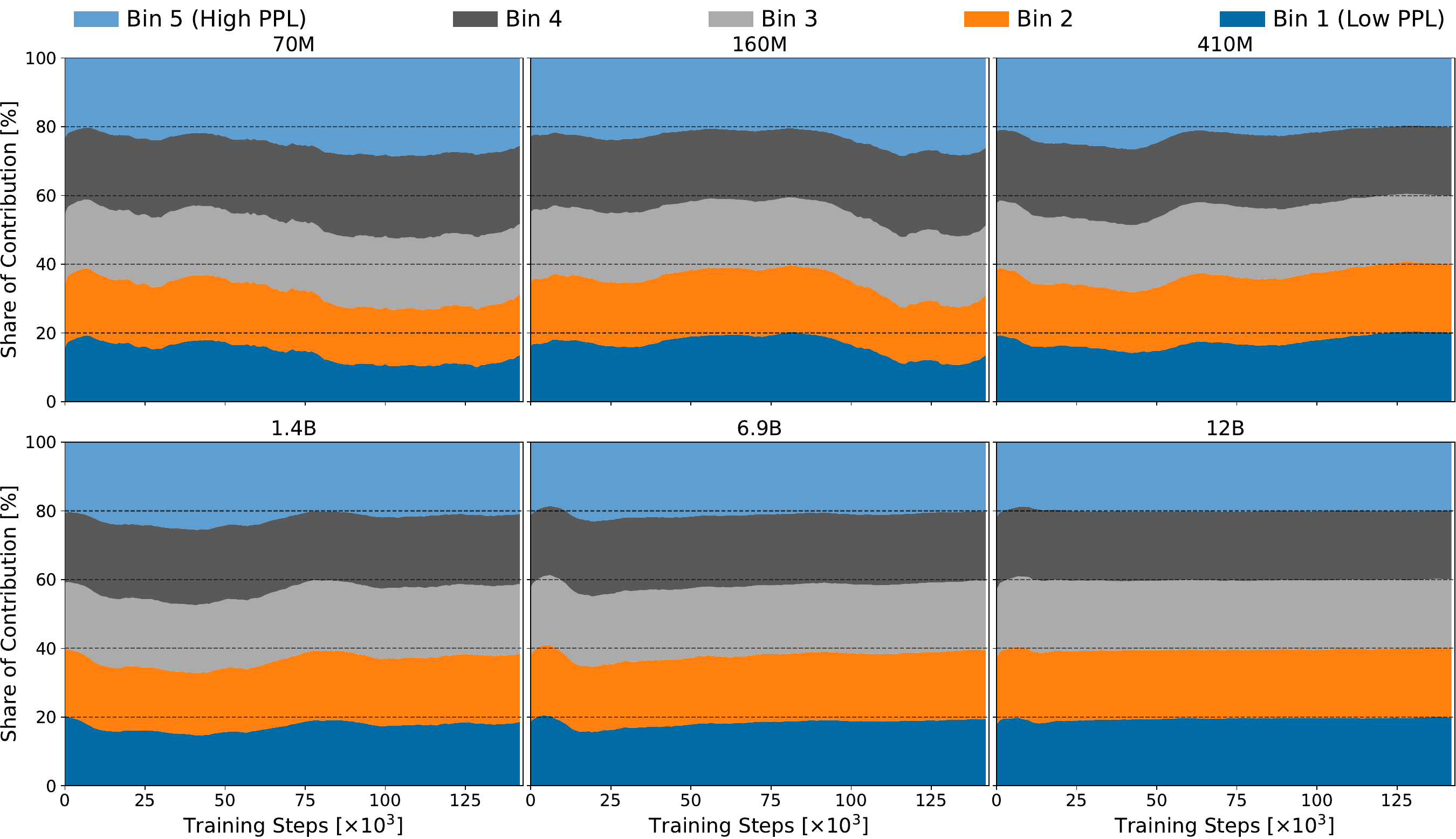}
  \caption{
  Relationship between text PPL and contribution across the six Pythia-Deduped model sizes.
  Each panel shows the share of normalized contribution attributed to five equal-sized PPL bins for one model size.
  All curves are moving averages with window size 10.
  }
  \label{fig:ppl_contribution_across_sizes}
\end{figure*}

\Cref{fig:ppl_contribution_across_sizes} extends the text-difficulty analysis in \cref{sec:contribution_difficulty}.
The 410M, 1.4B, and 6.9B models exhibit a highly similar pattern: higher-PPL examples account for a larger share of contribution during the middle stage of pretraining, while contribution is more evenly distributed across PPL bins during the early and late stages.
The 12B model shows the same tendency in the highest-PPL bin during the middle stage, although the magnitude of the shift is smaller than for the smaller models, consistent with the weaker contribution dynamics observed for larger models in our analysis.
By contrast, for the 70M and 160M models, the relative emphasis on higher-PPL examples emerges later, toward the end of training.
One possible interpretation is that smaller models progress through analogous training phases more slowly, causing this shift toward higher-PPL data to appear later in the pretraining trajectory.

\begin{figure*}[t]
  \centering
  \includegraphics[width=\textwidth]{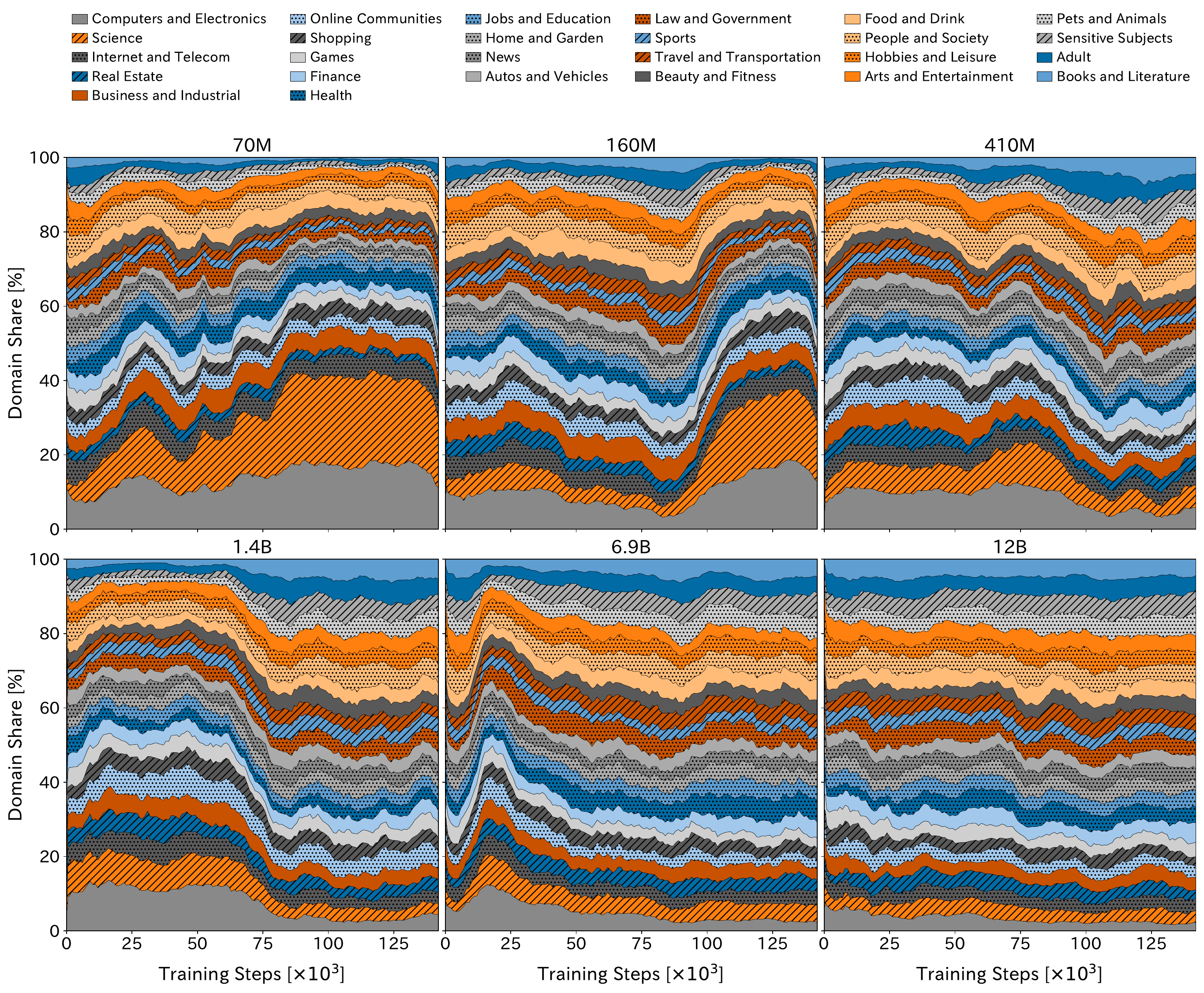}
  \caption{
  Domain composition of the bottom 5\% of contribution examples across the six Pythia-Deduped model sizes.
  Each panel corresponds to one model size.
  All curves are moving averages with window size 10.
  }
  \label{fig:bottom_domain_composition_across_sizes}
\end{figure*}

\begin{figure*}[t]
  \centering
  \includegraphics[width=\textwidth]{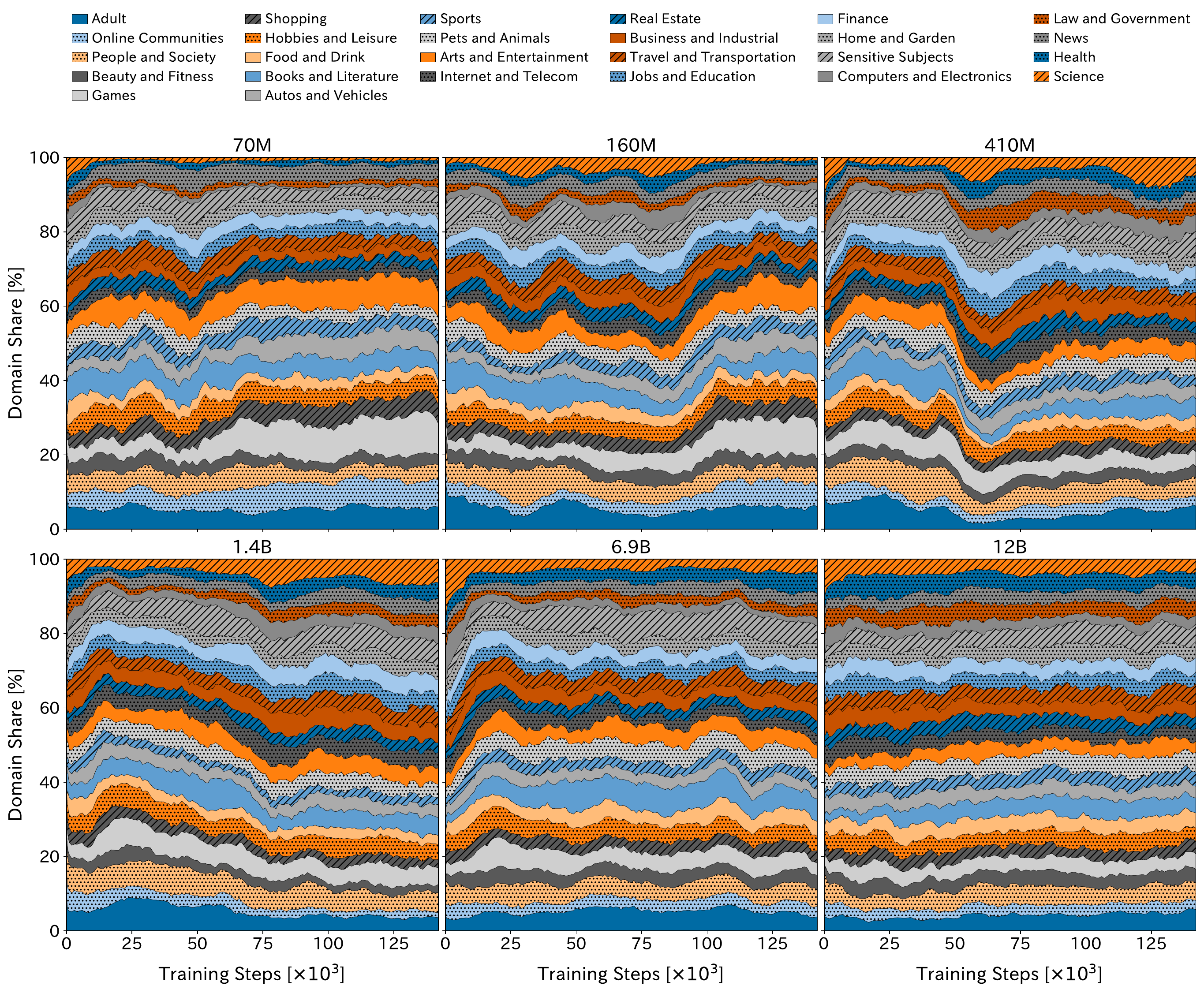}
  \caption{
  Domain composition of the top 5\% of contribution examples across the six Pythia-Deduped model sizes.
  Each panel corresponds to one model size.
  All curves are moving averages with window size 10.
  }
  \label{fig:top_domain_composition_across_sizes}
\end{figure*}

Finally, \cref{fig:bottom_domain_composition_across_sizes,fig:top_domain_composition_across_sizes} extend the domain analysis in \cref{sec:contribution_domain} across model sizes.
For the bottom 5\%, the 410M and larger models show broadly similar dynamics: STEM-related domains are prominent among low-contribution examples during the earlier stages and become less prominent later, while literature-related domains show the opposite tendency.
The magnitude of these changes tends to be smaller for larger models, consistent with the weaker contribution dynamics we observe at larger scales.
By contrast, for the 70M and 160M models, the share of STEM-related domains among low-contribution examples increases toward the end of training.
One possible interpretation, consistent with the delayed PPL dynamics discussed above, is that smaller models progress through analogous training phases more slowly.
For the larger models, the STEM share among low-contribution examples is initially small, increases during the first half of training, and then decreases later; the smaller models may primarily exhibit the earlier part of this progression within the observed training trajectory.
The top-5\% results are less uniform across model sizes.
The 410M and 1.4B models exhibit similar domain-level shifts, but we do not observe a single directional pattern that is shared across all six models.
Nevertheless, all model sizes exhibit clear phase-dependent changes in the domain composition of high-contribution examples.
Thus, while the specific domains that dominate the top contributors vary with model scale, the broader finding that domain-level contribution dynamics evolve across pretraining is preserved.

Taken together, the contribution-distribution, text-difficulty, and domain analyses reveal broadly consistent stage-dependent dynamics across model sizes, while also exhibiting systematic scale-dependent differences.
In particular, several transitions that occur during the middle stage for larger models appear later for smaller models, suggesting that the timing of training-phase transitions may depend on model scale.
We also observe that the magnitude of these dynamics tends to be smaller at larger scales.
Thus, model size appears to affect when and how strongly contribution patterns change across pretraining, while preserving much of their broader temporal structure.

\subsection{Consistency across Weight Initializations and Data Orderings}
\label{appendix:random_initialization}

We examine the robustness of the domain-level contribution dynamics to weight initialization and data ordering.
We first consider the standard PolyPythia runs at the 160M and 410M scales, using three available runs at each scale, corresponding to seeds 0, 1, and 3, in which both weight initialization and data ordering vary.\footnote{Seed 2 is omitted because the corresponding training data indices are missing from the released data at the time of writing.}
We then use the decoupled PolyPythia-160M variants to isolate each factor by varying either weight initialization or data ordering while holding the other fixed.

\begin{figure*}[t]
  \centering
  \includegraphics[width=\textwidth]{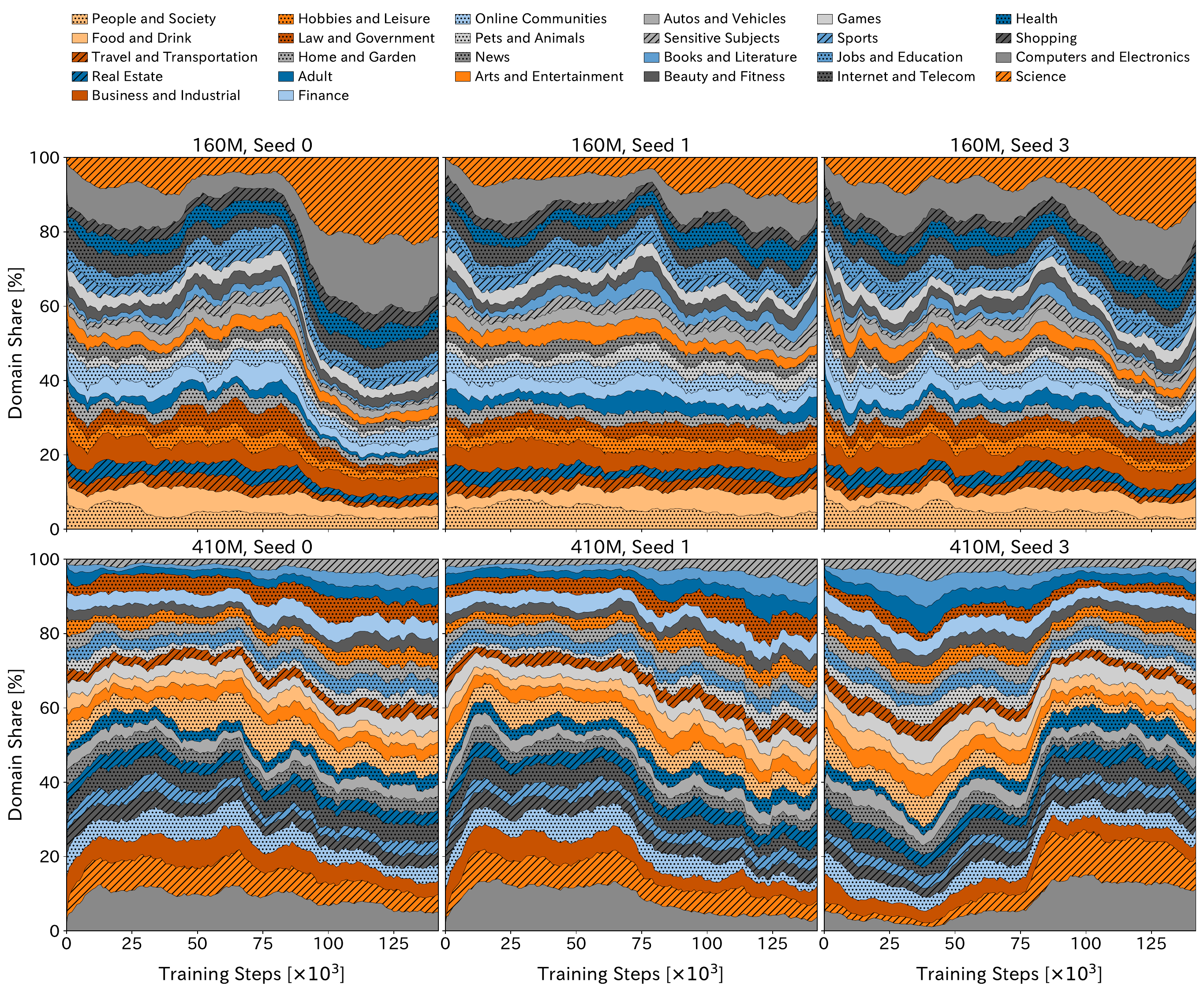}
  \caption{
  Domain composition of the bottom 5\% of contribution examples across standard PolyPythia-160M and PolyPythia-410M runs.
  Each model size includes three runs with different weight initializations and data orderings.
  All curves are moving averages with window size 10.
  }
  \label{fig:polypythia_joint_seed_bottom}
\end{figure*}

\begin{figure*}[t]
  \centering
  \includegraphics[width=\textwidth]{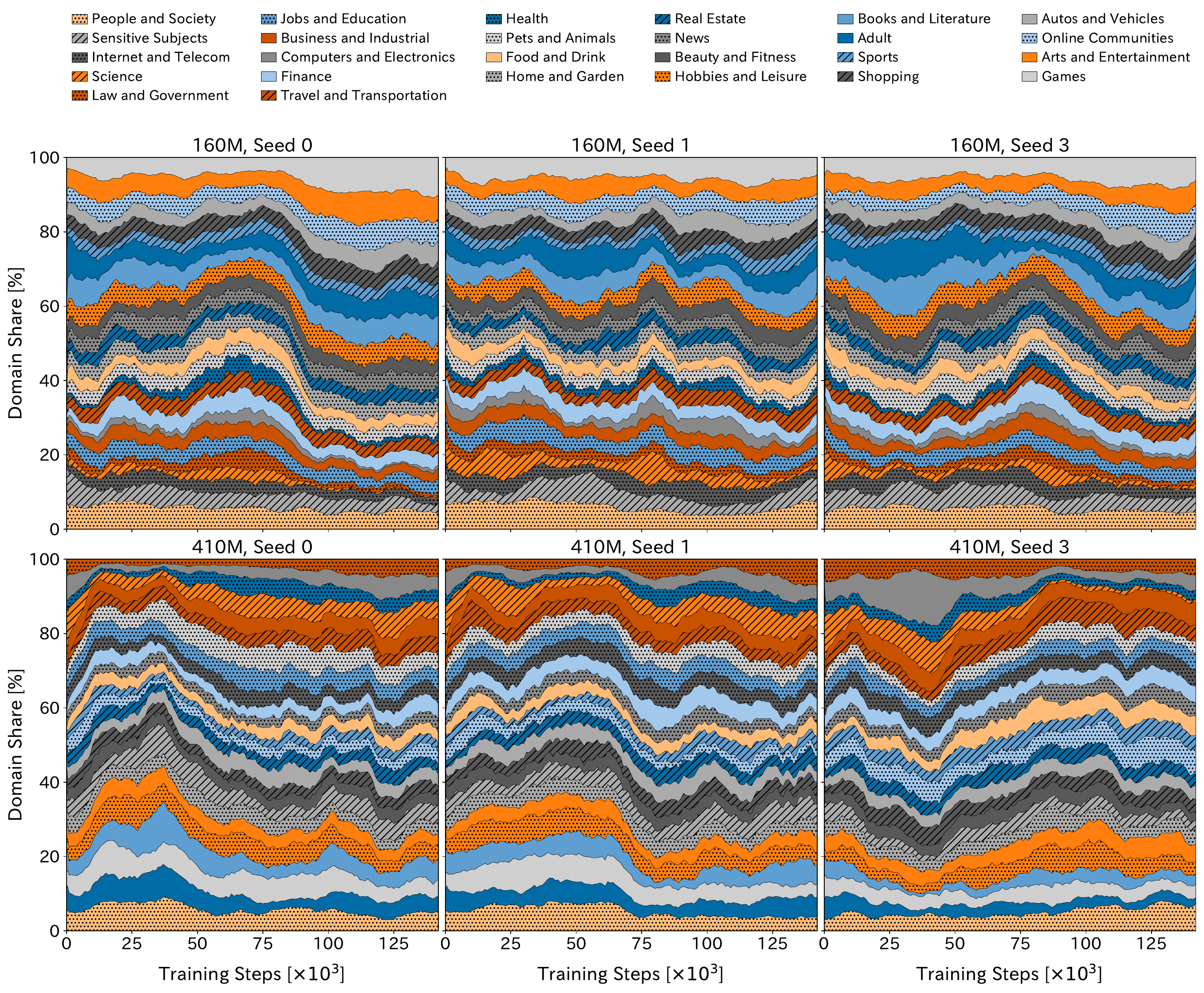}
  \caption{
  Domain composition of the top 5\% of contribution examples across standard PolyPythia-160M and PolyPythia-410M runs.
  Each model size includes three runs with different weight initializations and data orderings.
  All curves are moving averages with window size 10.
  }
  \label{fig:polypythia_joint_seed_top}
\end{figure*}

\Cref{fig:polypythia_joint_seed_bottom,fig:polypythia_joint_seed_top} show that the standard PolyPythia runs exhibit broadly consistent domain-level dynamics across seeds.
For the 160M runs, the share of STEM-related domains among low-contribution examples tends to increase as training progresses, whereas for the 410M runs, STEM-related domains are prominent early and become less prominent later.
The corresponding top-5\% patterns are likewise largely preserved across seeds, with similar domain-level transitions occurring over training.
An important exception is the 410M seed-3 run, which exhibits qualitatively different dynamics.
As reported by \citet{wal2025polypythias}, this run is an anomalous training run in which learning fails.
Its distinct contribution dynamics therefore appear to reflect the underlying training failure rather than ordinary seed-level variation.
Excluding this anomalous run, the broad temporal patterns are stable across the standard seed variants.

\begin{figure}[t]
  \centering
  \captionsetup[subfigure]{justification=centering, skip=2pt}

  \begin{subfigure}{\linewidth}
    \centering
    \includegraphics[width=\linewidth]{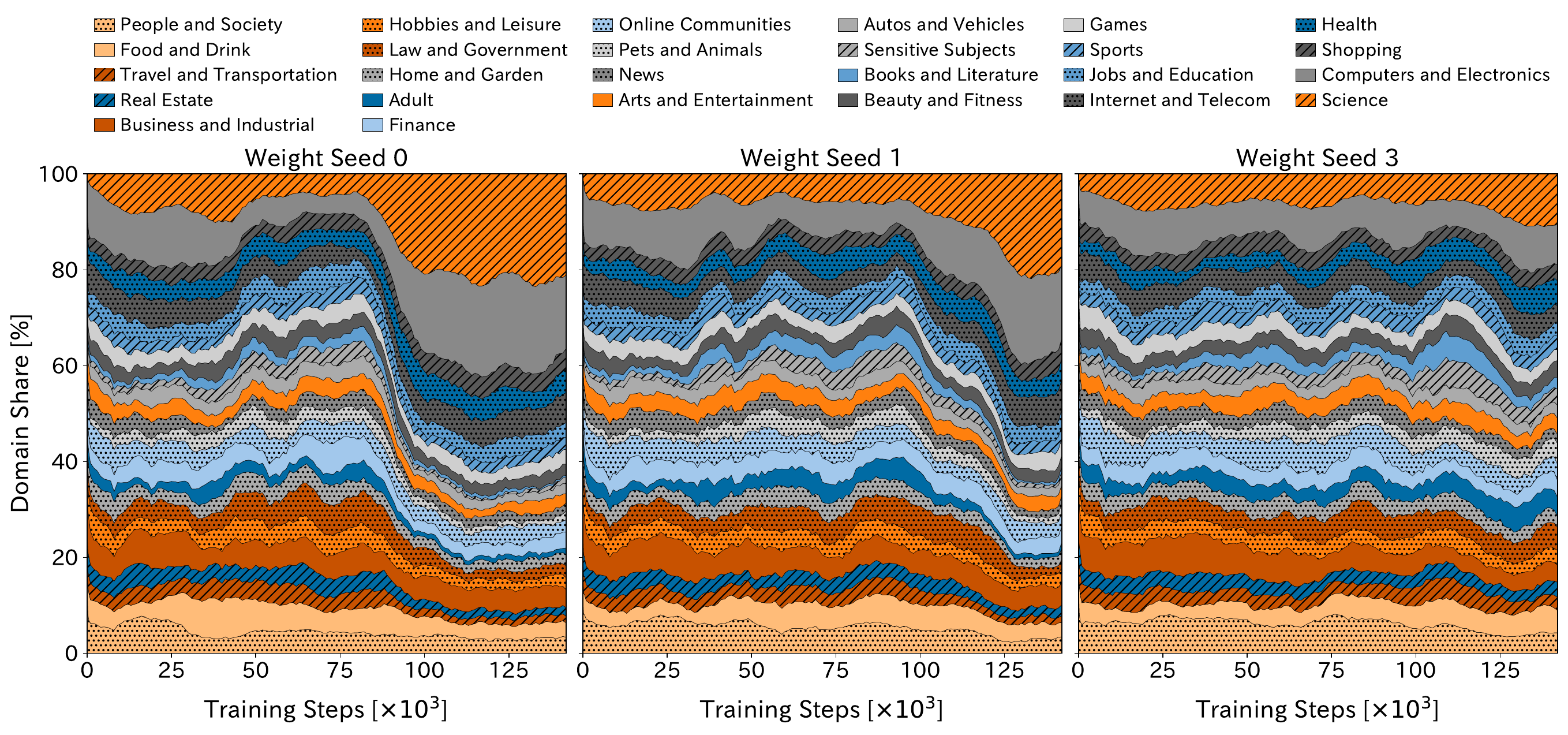}
    \caption{Bottom 5\% contribution examples, and their domain share in the overall data mix.}
    \label{fig:poly160m_weight_seed_bottom}
  \end{subfigure}

  \vspace{0.8em}

  \begin{subfigure}{\linewidth}
    \centering
    \includegraphics[width=\linewidth]{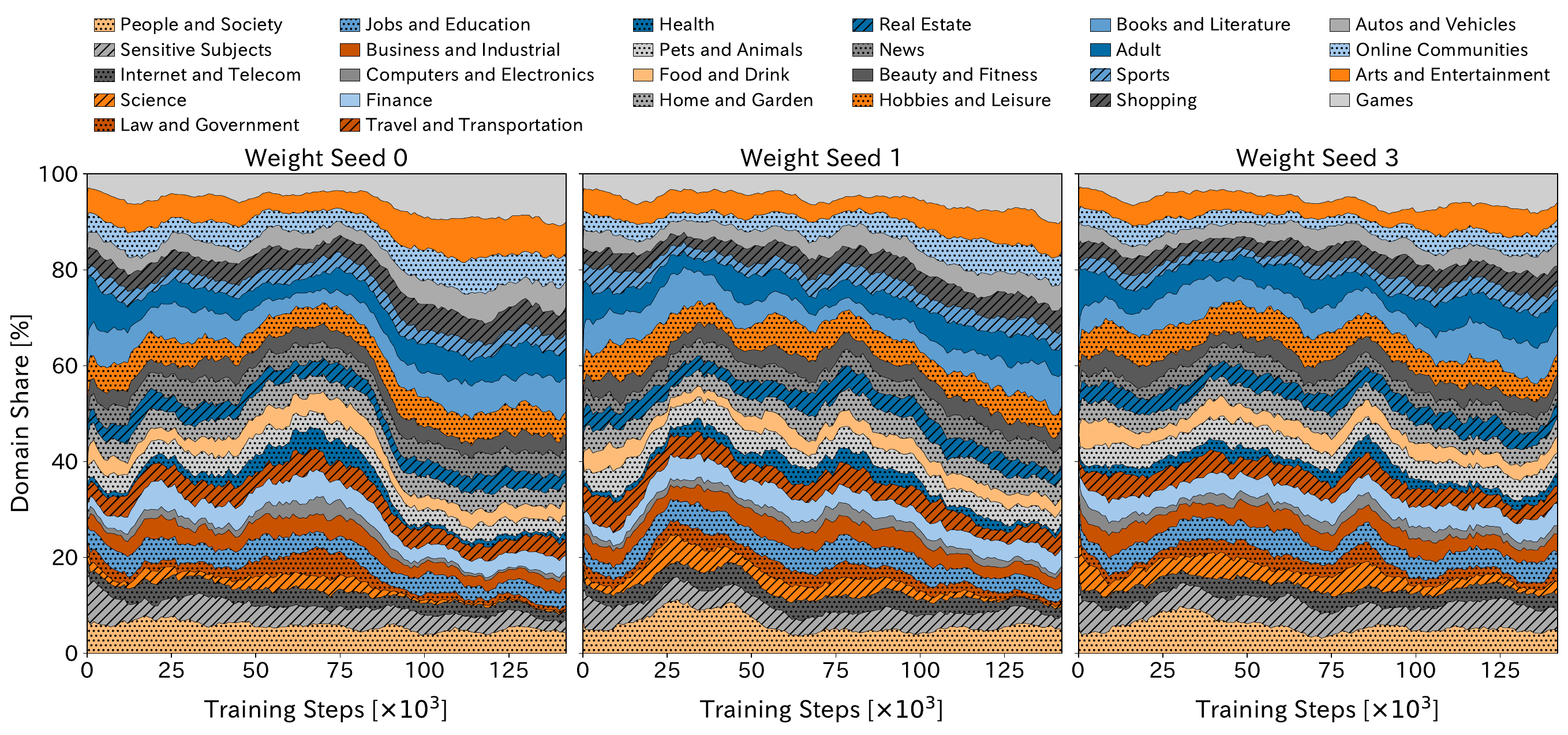}
    \caption{Top 5\% contribution examples, and their domain share in the overall data mix.}
    \label{fig:poly160m_weight_seed_top}
  \end{subfigure}

  \caption{
  Domain composition of extreme-contribution examples for PolyPythia-160M runs with different weight initializations.
  The qualitative contribution dynamics are highly consistent across runs when data ordering is fixed.
  }
  \label{fig:poly160m_weight_seed_domains}
\end{figure}

\begin{figure}[t]
  \centering
  \captionsetup[subfigure]{justification=centering, skip=2pt}

  \begin{subfigure}{\linewidth}
    \centering
    \includegraphics[width=\linewidth]{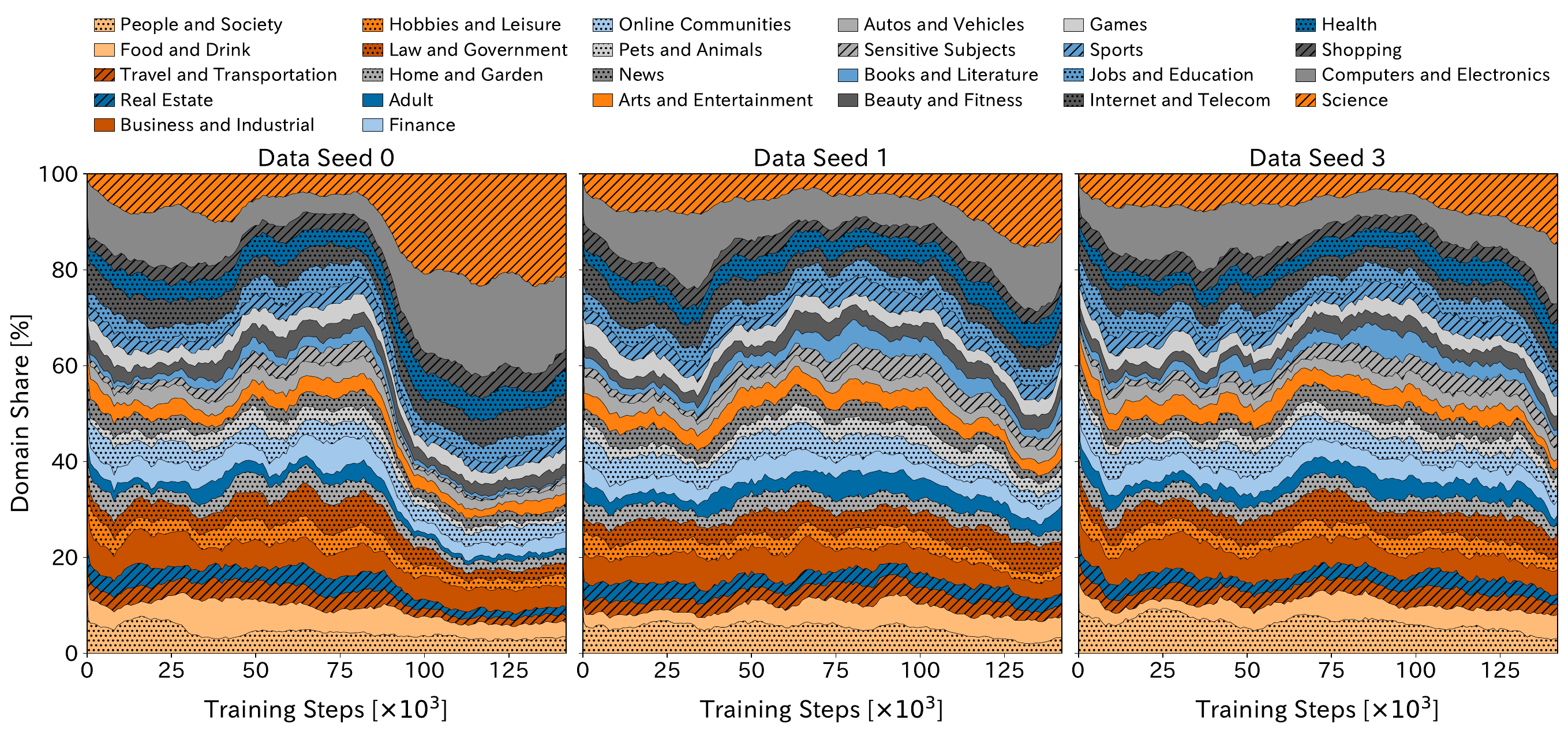}
    \caption{Bottom 5\% contribution examples, and their domain share in the overall data mix.}
    \label{fig:poly160m_data_seed_bottom}
  \end{subfigure}

  \vspace{0.8em}

  \begin{subfigure}{\linewidth}
    \centering
    \includegraphics[width=\linewidth]{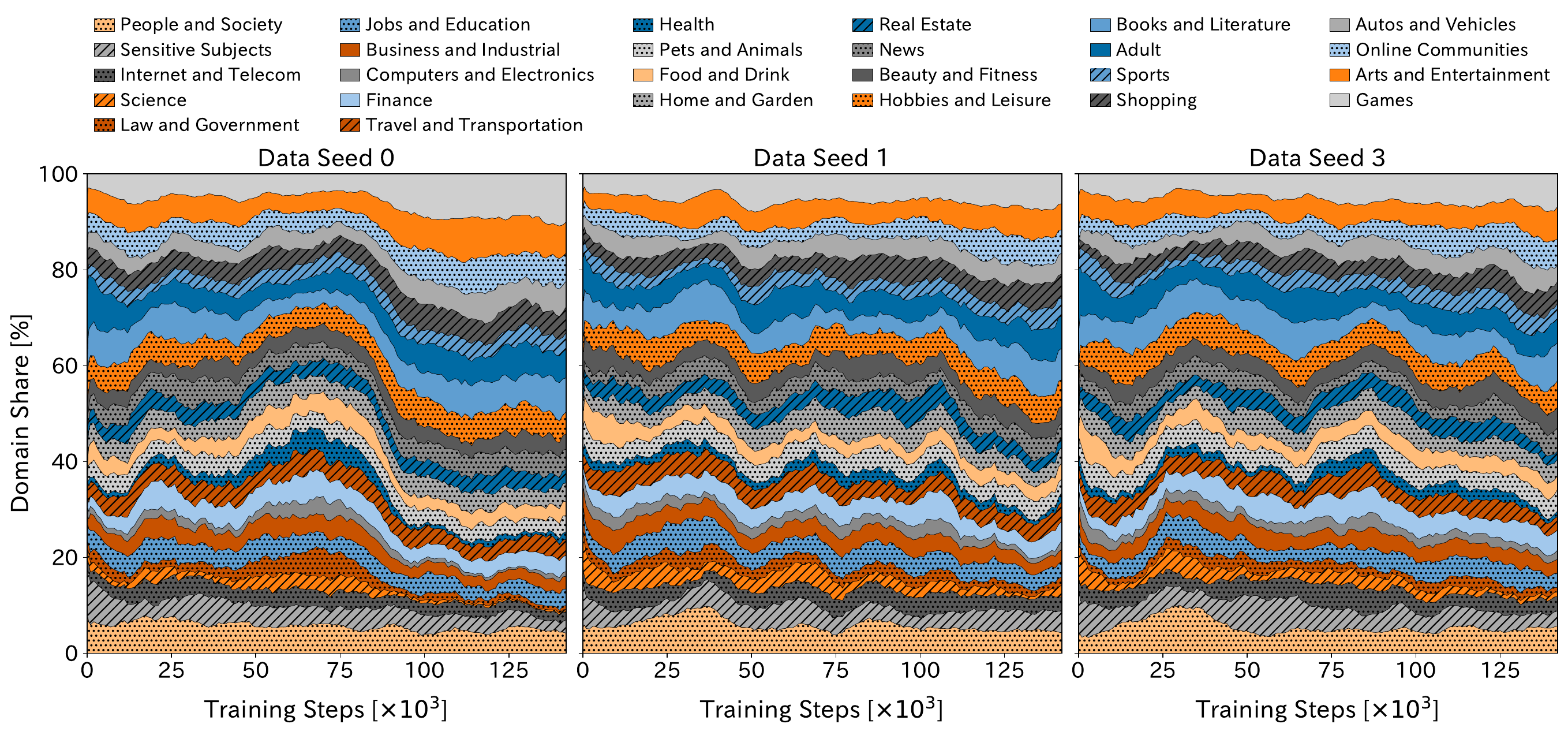}
    \caption{Top 5\% contribution examples, and their domain share in the overall data mix.}
    \label{fig:poly160m_data_seed_top}
  \end{subfigure}

  \caption{
  Domain composition of extreme-contribution examples for PolyPythia-160M runs with different data orderings.
  The broad temporal patterns remain similar across runs, although some variation is visible in the precise domain dynamics.
  }
  \label{fig:poly160m_data_seed_domains}
\end{figure}

To disentangle the effects of the two random factors, we next examine the decoupled PolyPythia-160M variants.
\Cref{fig:poly160m_weight_seed_domains} compares runs that differ only in weight initialization, while \cref{fig:poly160m_data_seed_domains} compares runs that differ only in data ordering.

When only weight initialization is varied, both the bottom- and top-contribution domain compositions are highly consistent across runs, with the same broad transitions occurring at similar stages of training.
Varying data ordering also preserves the overall temporal structure, although the bottom-5\% composition for seed 3 shows somewhat greater deviation from the other runs.
This suggests that the observed domain-level contribution dynamics may be somewhat more sensitive to data ordering than to weight initialization.

Taken together, these results indicate that the broad stage-dependent contribution dynamics are robust to ordinary seed variation, including independent changes in weight initialization and data ordering.
At the same time, the anomalous 410M seed-3 run shows that substantial training failures can produce qualitatively different contribution dynamics, while the decoupled variants suggest that data ordering may introduce somewhat greater variation than weight initialization.

\subsection{Details of the Task-Specific Influence Comparison}
\label{appendix:task_specific_comparison}

We provide the full experimental setup and results for the task-specific influence comparison summarized in \cref{sec:task_specific_comparison}.
For each validation domain $d$, let $\mathcal{V}_d$ denote its held-out validation set and define the corresponding average language-modeling loss at checkpoint $c$ as
\begin{equation}
L_d(\theta_c)
=
\frac{1}{|\mathcal{V}_d|}
\sum_{v \in \mathcal{V}_d}
\ell(v;\theta_c).
\end{equation}
For a training example $x$ processed within the checkpoint interval beginning at $c$, we compute the checkpoint-local TracIn-style score
\begin{equation}
\operatorname{TracIn}_d(x;c)
=
\nabla_\theta \ell(x;\theta_c)^\top
\nabla_\theta L_d(\theta_c).
\label{eq:tracin_domain_score}
\end{equation}
Up to the positive learning-rate factor shared within an interval, this gradient alignment approximates the first-order reduction in validation loss induced by an SGD update on $x$.
Because our analysis uses only within-interval rankings, we omit this common factor.

We construct the held-out validation sets from The Pile after removing examples that occur in the released Pythia pretraining stream.
We classify the remaining examples using the same NeMo Curator Domain Classifier as in \cref{sec:contribution_domain}.
For each of four domains, \textit{Books and Literature}, \textit{Science}, \textit{Computers and Electronics}, and \textit{Hobbies and Leisure}, we uniformly sample 1,000 examples assigned to that domain with classifier confidence greater than 0.95.
These four domains include both STEM and non-STEM validation targets and correspond to domains that exhibit prominent temporal patterns in our contribution analysis.

\begin{figure*}[t]
  \centering
  \captionsetup[subfigure]{justification=centering, skip=2pt}

  \begin{subfigure}{\textwidth}
    \centering
    \includegraphics[width=\textwidth]{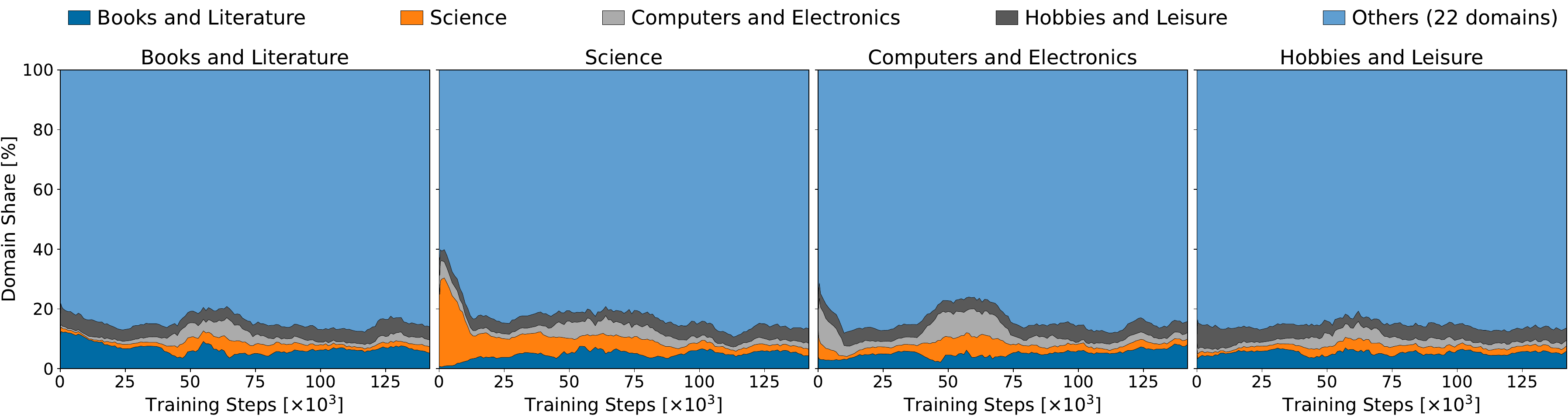}
    \caption{Domain composition of the top 5\% of training examples under the TracIn-style score.}
    \label{fig:tracin_top_domain_composition}
  \end{subfigure}

  \vspace{0.5em}

  \begin{subfigure}{\textwidth}
    \centering
    \includegraphics[width=\textwidth]{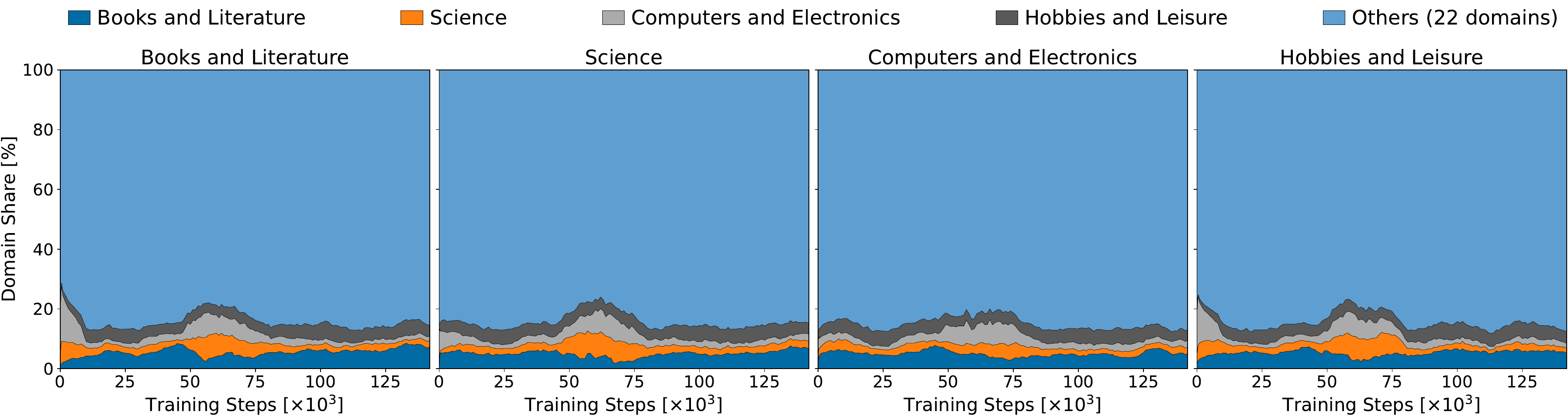}
    \caption{Domain composition of the bottom 5\% of training examples under the TracIn-style score.}
    \label{fig:tracin_bottom_domain_composition}
  \end{subfigure}

  \caption{
  Domain composition of training examples with extreme TracIn-style scores.
  Each panel uses the language-modeling loss on a different domain-specific held-out set as the validation target.
  Panel headings denote the validation domains used to compute the scores, while colors denote the source domains of the selected training examples.
  }
  \label{fig:tracin_domain_composition}
\end{figure*}

For each checkpoint interval of Pythia-1.4B-Deduped, we use the same domain-balanced pool of training examples as in \cref{sec:contribution_domain}.
Specifically, the pool contains 100 examples from each of the 26 domains.
We compute \cref{eq:tracin_domain_score} for every training example and each of the four validation domains, and then examine the domain composition of the top and bottom 5\% of examples under each score.
The curves in \cref{fig:tracin_domain_composition} are moving averages with a window size of 10 checkpoint intervals.

\end{document}